\documentclass[journal]{IEEEtran}
\usepackage{amsmath,amsfonts}
\usepackage{algorithmic}
\usepackage{algorithm}
\usepackage{array}
\usepackage[caption=false,font=normalsize,labelfont=sf,textfont=sf]{subfig}
\usepackage{textcomp}
\usepackage{stfloats}
\usepackage{url}
\usepackage{verbatim}
\usepackage{graphicx}
\usepackage{cite}
\usepackage{wrapfig}

\begin{document}


\title{Deep Learning Approaches in \\Pavement Distress Identification: A Review}

\author{Sizhe~Guan, Haolan~Liu, Hamid~R.~Pourreza, and~Hamidreza~Mahyar*
\thanks{* Corresponding author: Hamidreza Mahyar (\texttt{mahyarh@mcmaster.ca}).}
\thanks{S. Guan, H. Liu, and H. Mahyar are with the Faculty of Engineering, McMaster University, Ontario, Canada.}
\thanks{H. Pourreza is with the Faculty of Engineering, Ferdowsi University of Mashhad, Iran.}
\thanks{Manuscript received April 19, 2005; revised August 26, 2015.}}

\markboth{Journal of \LaTeX\ Class Files,~Vol.~14, No.~8, August~2021}%
{Guan \MakeLowercase{\textit{et al.}}: Deep Learning Approaches in Pavement Distress Identification: A Review}


\maketitle

\begin{abstract}
This paper presents a comprehensive review of recent advancements in image processing and deep learning techniques for pavement distress detection and classification, a critical aspect in modern pavement management systems. The conventional manual inspection process conducted by human experts is gradually being superseded by automated solutions, leveraging machine learning and deep learning algorithms to enhance efficiency and accuracy. The ability of these algorithms to discern patterns and make predictions based on extensive datasets has revolutionized the domain of pavement distress identification.
The paper investigates the integration of unmanned aerial vehicles (UAVs) for data collection, offering unique advantages such as aerial perspectives and efficient coverage of large areas. By capturing high-resolution images, UAVs provide valuable data that can be processed using deep learning algorithms to detect and classify various pavement distresses effectively.
While the primary focus is on 2D image processing, the paper also acknowledges the challenges associated with 3D images, such as sensor limitations and computational requirements. Understanding these challenges is crucial for further advancements in the field.
The findings of this review significantly contribute to the evolution of pavement distress detection, fostering the development of efficient pavement management systems. As automated approaches continue to mature, the implementation of deep learning techniques holds great promise in ensuring safer and more durable road infrastructure for the benefit of society.
\end{abstract}


\section{Introduction}
\IEEEPARstart{P}{avements} hold a crucial societal role, exerting significant influence on both the economy and transportation infrastructure. The escalating costs associated with road maintenance and rehabilitation endeavors have underscored the utmost importance of pavement condition detection. This detection process encompasses four principal measurement units: Pavement Condition Index (PCI) for discerning surface distresses, International Roughness Index (IRI) for evaluating road roughness, Structural Number (SN) for assessing structural adequacy, and Skid Resistance (SR) \cite{SHOLEVAR2022104190} for quantifying safety aspects. Limited financial resources frequently engender delays in road maintenance services, exacerbating the challenges faced in preserving pavement quality. Moreover, the combined effects of heightened traffic loads and capricious climatic and environmental conditions expedite pavement deterioration, precipitating an exponential surge in maintenance costs. Failure to execute timely maintenance strategies throughout the pavement's projected service life culminates in structural deficiencies, mandating extensive rehabilitation or even complete reconstruction efforts, which prove financially unfeasible\cite{peraka2020pavement}. The manual detection of pavement distresses is labor-intensive and susceptible to the inadvertent oversight of minor anomalies due to inherent limitations in human visual inspection capabilities.

With the introduction of Artificial Intelligence (AI) algorithms into the field of pavement detection, significant advancements have been achieved \cite{doi:10.1061/JPEODX.0000373}. Xu and Zhang \cite{doi:10.1061/JPEODX.0000373} provided a comprehensive overview of how AI algorithms have been utilized in various stages of pavement management within the period of 2015 to 2020. They outlined the application of AI algorithms in the pavement management cycle and categorized them into three types: distress evaluation, performance modeling, and M\&R programming. In this paper, our emphasis lies on the recent implementation of deep learning algorithms for pavement distress evaluation, particularly on UAV-based 2D images. The entire process is divided into three main sections:

\begin{itemize}
  \item Data Acquisition and Preparation
  \item Distress Detection and classification 
  \item Model Evaluation
\end{itemize}

To enhance the quality of datasets during the image acquisition process for pavement distress detection, the integration of AI and Internet of Things (IoT) based techniques can offer notable advantages. These techniques alleviate the subjectivity inherent in manual measurements and surmount limitations arising from geographical location and adverse weather conditions during data collection. In the specific context of utilizing two-dimensional (2D) images for pavement distress detection, image-based data acquisition primarily entails the use of smartphones, cameras, and unmanned aerial vehicles (UAVs)\cite{SHOLEVAR2022104190}. Smartphones and cameras can be employed either through manual capturing of photos or by installing devices on manned vehicles to acquire images while in motion. However, UAVs exhibit superior flexibility and mobility, enabling the execution of tasks that are arduous to accomplish manually \cite{zakeri2017image}.

Pavement distress analysis entails two primary approaches: traditional image processing methods and learning-based algorithms. Traditional image processing techniques extract low-level distress features and employ them for distress detection and classification. In contrast, learning-based algorithms, such as machine learning (ML), leverage their computational power and autonomous learning capabilities to uncover hidden features and rules within pavement distresses, thus facilitating more accurate detection compared to human visual inspection  \cite{doi:10.1061/JPEODX.0000373}.

This paper aims to provide a comprehensive review of the current state-of-the-art in AI applications for pavement distress detection and classification. Additionally, it seeks to shed light on the utilization of unmanned aerial vehicles (UAVs) for acquiring pavement distress images, thereby offering valuable insights into the potential of UAVs and AI methods for future research in this domain.

Considering the limited availability of published literature specifically focused on aerial images, this review explores and introduces methods that could potentially be applied to such imagery. The paper's structure encompasses various aspects: the initial section addresses UAV-based image acquisition methods, followed by discussions on distress detection algorithms, public datasets, system evaluation techniques, and the availability of open-source codes. Finally, the paper concludes by discussing potential avenues for future research in this field.
\begin{figure*}[t]
\begin{center}
  \includegraphics[width=0.8\textwidth]{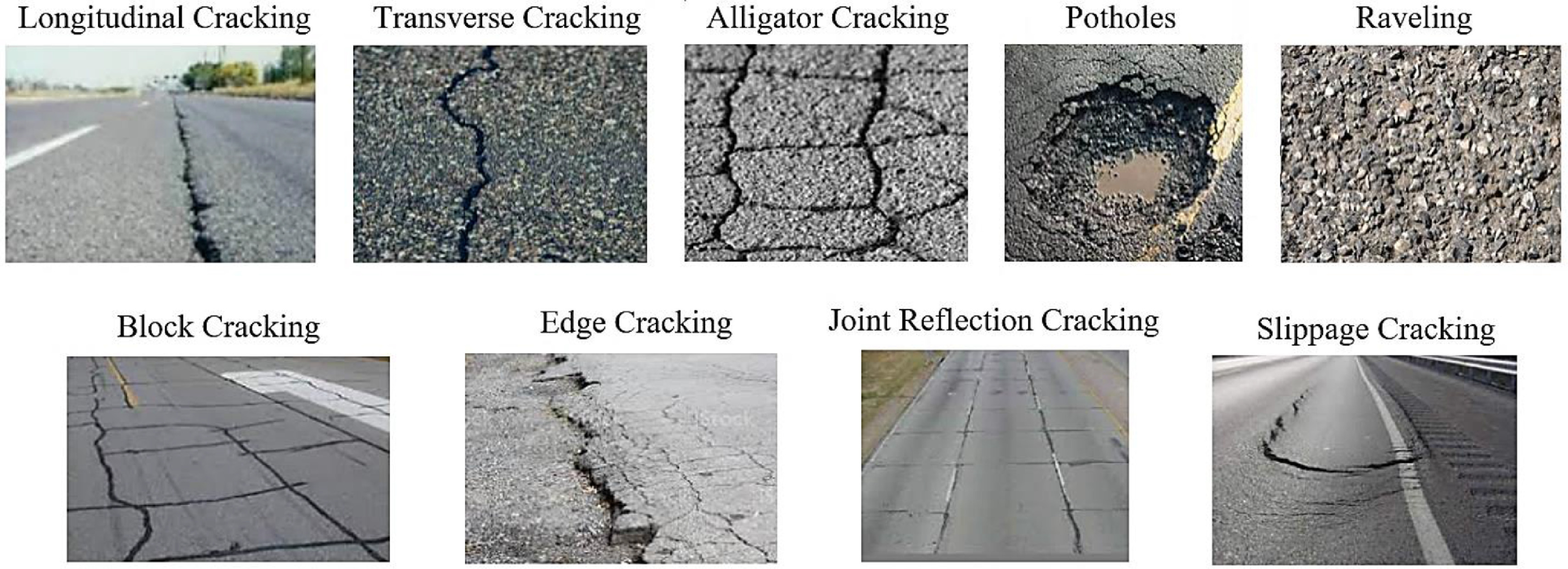}
  \caption{Multiple types of road pavement failures.\cite{khawaja2018review, Pothole, Kim_2021a, Bradshaw_2020, PPRA, Indiana_2021_IDEA_Block_Cracking, IMS, Mallela_Weiss_2009, coleri2017hmac}}
  \label{fig1}
\end{center}
\end{figure*}

\section{Pavement distress categorization}
Pavement distress encompasses various types, including patching, cracks, miscellaneous distress, pavement deformation, and defects \cite{RAJENDRA2015285}. Of these, cracks hold particular significance as they constitute a prevalent form of distress that has attracted considerable attention in prior investigations due to their widespread occurrence. The classification of crack types hinges on their length, depth, and width, giving rise to several common categories: transverse cracks, longitudinal cracks, block cracks, alligator cracks, raveling, edge cracks, reflection cracks, slippage cracks, and potholes.

Transverse cracks are fissures that traverse the material or object, exhibiting a perpendicular orientation to its center line. Conversely, longitudinal cracks occur parallel to the pavement's center line and are often linked to factors such as thermal expansion, contraction, or heavy loads. Block cracks manifest as interconnected fissures forming square or rectangular patterns arising from the shrinkage of the asphalt binder. Alligator cracks present an interconnecting or interlaced pattern within the asphalt layer, resembling the hide of a crocodile \cite{enwiki:1149726524}. Raveling pertains to dislodging coarse aggregate on the pavement surface \cite{WinNT}. Edge cracks, situated in parallel and typically within 0.3 to 0.5 meters (1 to 1.5 feet) of the pavement's outer edge, emanate from insufficient support, heavy traffic, or moisture infiltration \cite{manual1997pavement}. Reflection cracks emerge in the overlay or surface course of pavement as a consequence of movement within the rigid pavements beneath the asphalt concrete (AC) surface due to thermal and moisture variations. Slippage cracks adopt a crescent-shaped pattern and arise from a low-strength surface mix or inadequate bond between the surface and the underlying pavement layer, leading to surface deformation when wheels brake or turn. Potholes depict bowl-shaped depressions on the pavement surface from water infiltration into the top asphalt layer through road cracks. Fig. 1 illustrates various types of road pavement failures.

\section{UAV-based image acquisition}
The techniques used for data acquisition in pavement distress analysis can be categorized into manual (human-based), automated (machine-based), and semi-automated (a combination of human and machine) approaches \cite{abou2017fully}. Fig. 2 provides a visual overview of the pavement image acquisition process and shows different techniques employed in data acquisition.

\begin{figure*}[t]
\begin{center}
  \includegraphics[width=0.9\textwidth]{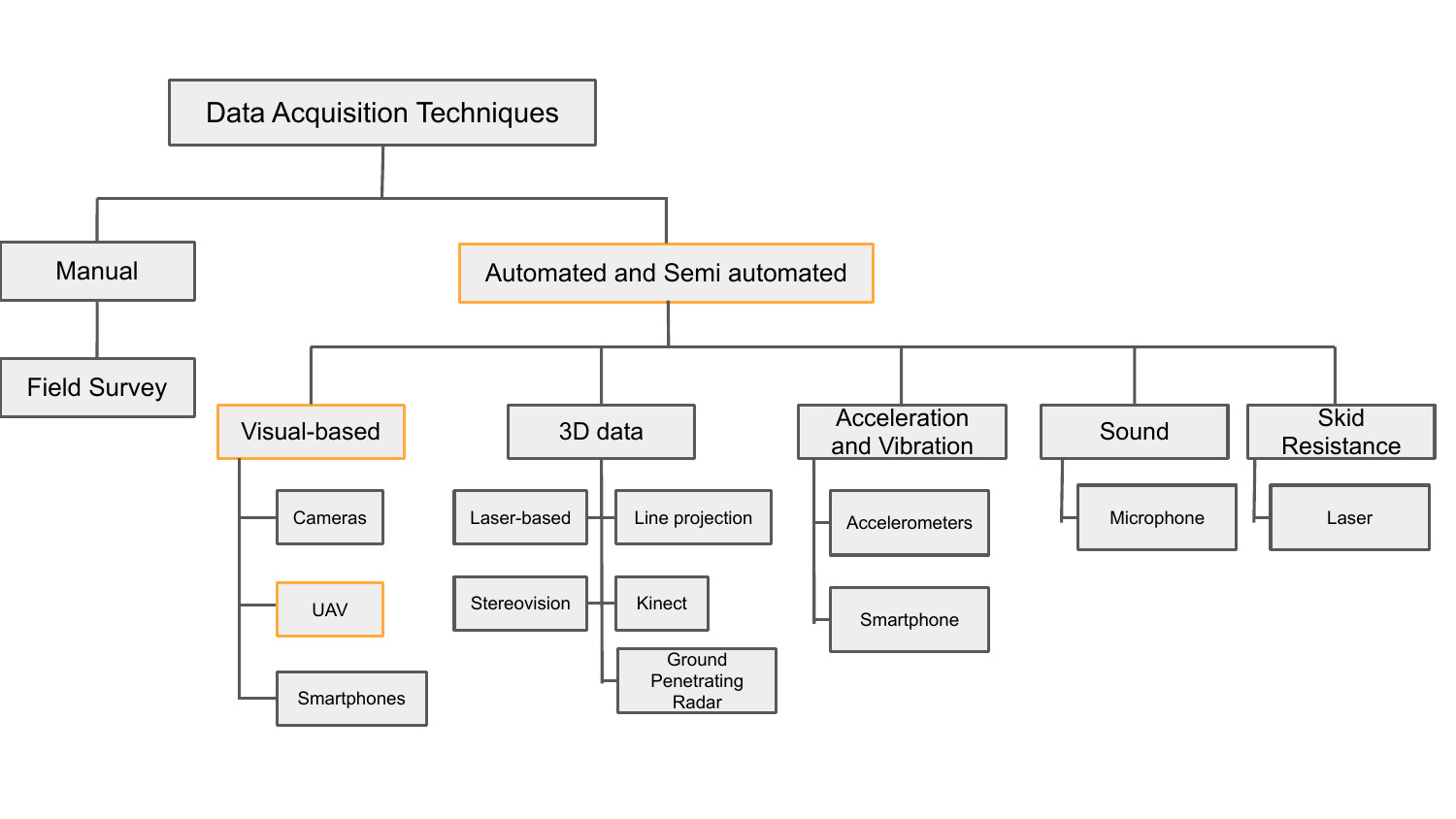}
  \vspace{-25pt}
  \caption{Data collection techniques. The Orange boxes demonstrate the data acquisition techniques focused in this review paper}
  \label{fig2}
\end{center}
\end{figure*}

Manual data collection is often regarded as inefficient and costly \cite{SHOLEVAR2022104190}, and the datasets obtained through this approach may present challenges in terms of processing for machine learning. In contrast, automated and semi-automated data collection techniques leverage various sensors, including smartphones, cameras, and unmanned aerial vehicles (UAVs).

Visual-based approaches involve the utilization of digital cameras, either by directly capturing photos or by integrating them into different devices, such as vehicles and UAVs, to enhance the adaptability of image acquisition. This approach finds extensive application in pavement distress detection and classification. Notably, the advantage of the visual-based approach lies in the swift processing of its 2D images using contemporary ML techniques, while datasets obtained through alternative means may necessitate more intricate preparations \cite{SHOLEVAR2022104190}.

In addition to 2D image data, 3D data is also commonly employed in pavement distress detection and classification. Compared to 2D image data acquired through visual-based approaches, 3D data provides more comprehensive information and holds the potential to improve AI model performance. Ground Penetrating Radar (GPR) is a notable technology capable of detecting concealed distresses that might be challenging to discern in 2D images. For instance, Liang et al. \cite{abou2017fully} utilized VGG16 and ResNet50 models to analyze 3D-GPR pavement distress images, yielding high computational efficiency and an average classification accuracy of 0.9 on the pavement distress classification task \cite{liang2022automatic}.

Acceleration and vibration data analysis rely on measuring vertical differences induced by pavement distress, facilitating the detection of anomalies such as potholes. This approach involves monitoring the pressure and vibration of the tire to discern distresses' presence. Utilizing an accelerometer is a prevalent method to detect distress, as it captures the relative movements of a vehicle in three dimensions. When the vehicle traverses distresses, causing vertical differences, the shock experienced by the vehicle allows for the detection of these anomalies.

Furthermore, distress can be detected by monitoring tire-generated sound, which can be captured through strategically positioned microphones near the wheel. Analyzing tire sound permits the detection of distresses such as raveling and bleeding, in addition to those causing pronounced vertical differences \cite{coenen2017review}. These alternative methods of distress detection contribute valuable insights and can complement visual-based approaches by providing diverse data types for analysis.

The UAV system, characterized by its rapid speed, high maneuverability, and cost-effectiveness for 2D image data collection, has exhibited substantial potential to emerge as a primary method for pavement detection in the future \cite{zakeri2017image}. The capability of drones to transport lightweight equipment renders UAVs low-cost data collection platforms with remarkable maneuverability \cite{SHOLEVAR2022104190}. Employing UAVs in image acquisition furnishes a valuable resource of images well-suited for detecting distresses such as cracks, potholes, and rutting \cite{SHOLEVAR2022104190}. UAVs offer the distinct advantage of capturing expansive images with acceptable resolution, enabling the creation of datasets containing sufficient images for training AI models in a relatively brief timeframe.

Kim et al. \cite{kim2017concrete} developed a UAV-based prototype and applied it to acquire images for crack identification. The efficacy of digital image processing algorithms is contingent upon the quality of images obtained through UAVs. Although UAVs offer an efficient means to acquire high-quality 2D image datasets for pavement distress, the process of UAV image acquisition faces challenges related to image degradation. Notably, blurring often arises due to the movements and vibrations of UAVs \cite{liu2020deep}.

To address this issue, Zhu et al. \cite{zhu2022pavement} proposed a method for calculating optimal UAV flight settings to ensure high image quality. The tuning process involves adjusting flight parameters and camera parameters based on the desired image resolution. Specifically, the ratio of the focal camera sensor size (mm) to the pavement width (m) should be equivalent to the ratio of the focal length of the camera (mm) to the flight altitude (m). Additionally, flight speed should be considered to prevent excessive overlaps between captured images. The flight speed (v) can be determined using the equation tl(1-r), where t represents the sampling frequency (Hz), l denotes the actual road segment length captured in one image, and r represents the frontal overlap \cite{zhu2022pavement}. By employing these tuned parameters, the UAV can acquire images with higher quality based on the designated flight plan.
In addition, Mao, Z. et al. \cite{mao2020research} have developed a pavement disease recognition and perception system based on the application UAV, consisting of four layers:

\begin{itemize}
  \item Pavement disease sensing layer
  \item Disease data storage layer 
  \item Disease data processing layer
  \item System user layer
\end{itemize}

According to the findings of Mao et al. \cite{mao2020research}, unmanned aerial vehicles (UAVs) exhibit superior flexibility and efficiency in the domain of pavement disease detection compared to conventional road disease detection vehicles. These studies underscore the immense potential of UAVs in capturing high-quality images for pavement distress detection and classification. Moreover, the optimization of flight settings emerges as a crucial factor in enhancing image quality, emphasizing the significance of meticulous planning and parameter adjustment to maximize the utility of UAV-based data acquisition.
\section{Distress detection and classification algorithms}
Distress detection and classification algorithms can be classified into two main categories: 1) traditional methods, where feature extraction is carried out manually, and 2) learning-based methods, where feature vectors are generated through a process of learning. Detailed explanations of these two approaches will be provided in the subsequent sections.
\subsection{Traditional algorithms}
Traditional algorithms typically comprise several components (Fig. 3) designed to process raw image data and extract valuable information from input images. The primary components of a traditional algorithm include the following:
\begin{enumerate}
  \item Preprocessing: This component involves manipulating and enhancing the raw image data.
  \item Feature extraction: This component involves identifying and extracting specific features or structures from the image data, such as edges, corners, or shapes. Feature extraction may involve techniques such as thresholding, filtering, or segmentation.
  \item Feature representation: This component involves representing the extracted features in a suitable format for further processing or analysis. For example, features may be represented as numerical vectors or graphs.
  \item Classification: This component involves assigning a label or category to the image data based on the extracted features. Classification may involve techniques such as supervised learning, unsupervised learning, or template matching.
\end{enumerate}

\begin{figure}
  \includegraphics[width=\linewidth]{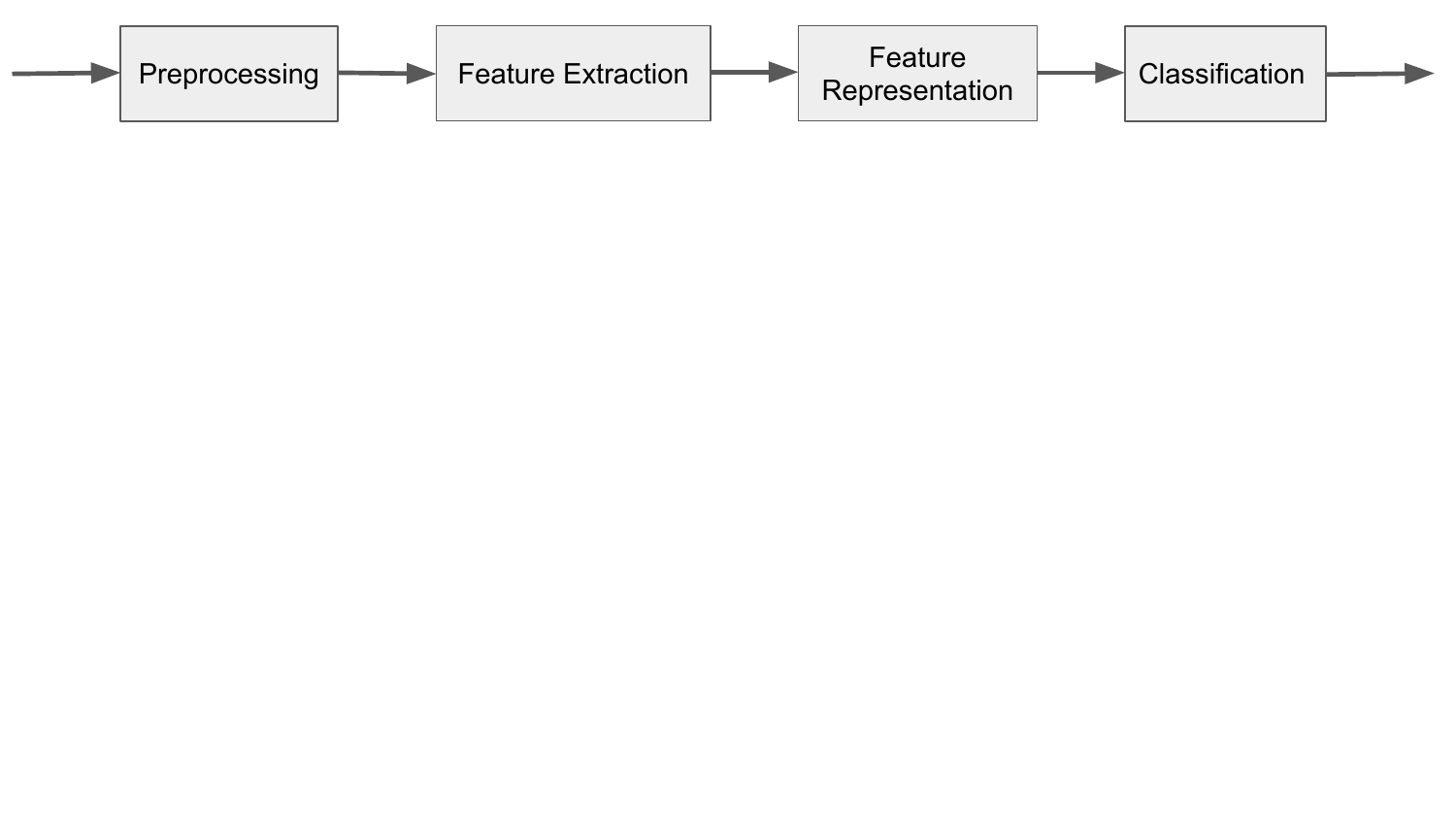}
  \vspace{-130pt}
  \caption{Components in traditional algorithms.}
  \label{fig3}
\end{figure}
\bigskip

    \subsubsection{\textbf{Preprocessing}}
   Preprocessing plays a critical role in image processing as it improves the quality and suitability of the image data for further analysis or interpretation.
   \vspace{3pt}

   \textbf{Linear filters}: The initial prominent image processing technique applied involves a linear smoothing filter. A linear filter is characterized as a process wherein the filter's transfer function alters a portion of the signal frequency spectrum. Typically, these filters exhibit linearity and shift-invariance. Consequently, the output images are obtained through convolving the input image with the filter's impulse response. This technique aims to mitigate intensity fluctuations in regions arising from disruptive influences, such as noise, or intrinsic image characteristics like edges and fine textures \cite{zhang2013matched}.
\vspace{3pt}

       \textbf{Gaussian filters}: The Gaussian filtering approach involves replacing each pixel value in an image with the weighted mean of its neighboring pixels.  The weights of the kernel are determined based on the Gaussian distribution. The following function represents the zero-mean distribution with variance $\sigma^2$ :
       \[ h(i,j) =ke^{-\frac{i^2 + j^2}{2\sigma^2}} \]
       The variance parameter $\sigma$ controls the level of smoothing applied by the filter and k is the constant that normalizes the equation.
       
       \begin{figure}
       \begin{center}
        \includegraphics[width=0.8\linewidth]{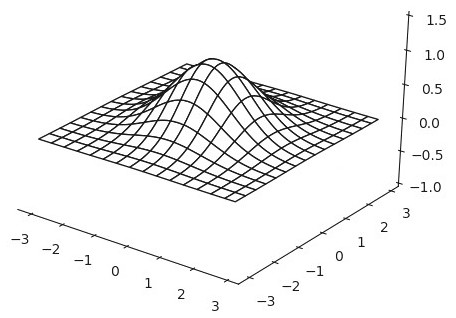}
        \caption{2D Gaussian with zero mean and $\sigma$=1.\cite{Gramaglia_2020}}
        \label{fig4}
        \end{center}
        \end{figure}
        
        Fig. 4 shows a 2D Gaussian with zero mean and $\sigma$=1 \cite{Gramaglia_2020}. Gaussian filters are preferred over other filters due to their consistent smoothing strength across all directions. This means that Gaussian filters are effective regardless of the image's orientation. They excel in preserving local features by seamlessly blending the processed patches with the surrounding areas, resulting in a more natural appearance.
       \vspace{3pt}
       
       \textbf{Mean filters}: The mean filtering technique entails the substitution of each pixel value in an image with the average value of its neighboring pixels. The initial step involves defining a neighborhood around the pixel, usually accomplished by employing a square window of a predetermined size, such as 3x3 or 5x5. Subsequently, the mean value is computed as the average of all pixels within the specified neighborhood. Following this, the current pixel's value is replaced with the computed mean value. This procedure is iterated for all pixels in the image, culminating in a smoothed image with diminished noise.
\vspace{3pt}

    \textbf{Nonlinear filters}:
         Non-linear filters are a category of image processing filters that serve to enhance or manipulate image features. Unlike their linear counterparts, which apply straightforward arithmetic operations on pixel values within a neighborhood, non-linear filters employ more intricate algorithms capable of executing operations that cannot be readily characterized by linear functions. These filters demonstrate particular efficacy in noise reduction and artifact elimination from images, while also being adept at enhancing specific features or structures within an image.  
 \vspace{3pt}

       \textbf{Morphological filters}: 
Morphological filtering is an image processing technique that considers the shapes and relationships between objects, rather than relying solely on pixel values, in its operations. Unlike linear filters, morphological filtering does not involve a coefficient matrix; instead, it employs a matrix to define a structure element (SE) with values ranging from 0 to 1. The size and shape of the SE can be adjusted to cater to the specific requirements of the image-processing task. Morphological filtering is based on two fundamental operations: dilation and erosion. Dilation, as a morphological operator, facilitates the growth process by expanding the size of an object in an image through the addition of pixels to its boundaries. Conversely, erosion, the opposite morphological operator, facilitates the shrinkage process by reducing the size of an object in an image through the removal of pixels from its boundaries. The shape of the SE influences the speed and direction of both growth and shrinkage processes. Figure 5 illustrates the processes of shrinking and growing \cite{Thipkham_2019}.
       
       \begin{figure}[t]
       \begin{center}
        \includegraphics[width=0.9\linewidth]{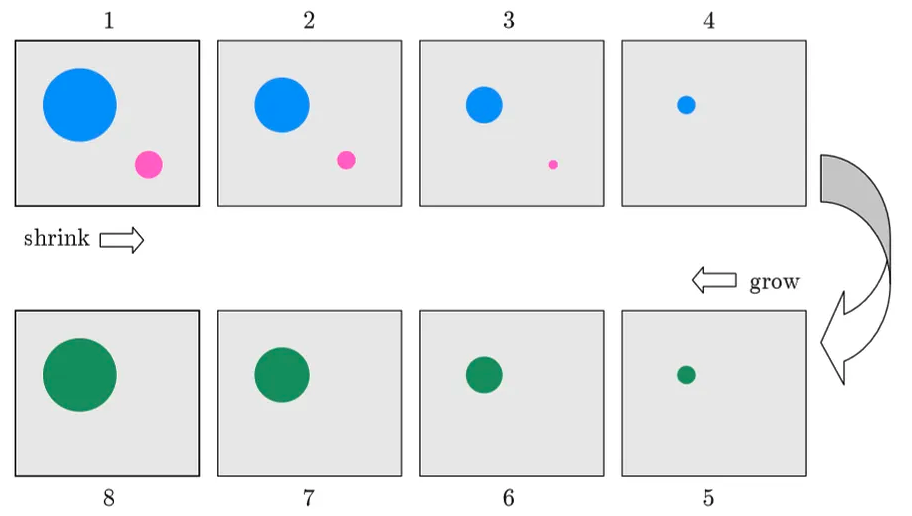}
        \caption{Shrinking and growing processes. \cite{Thipkham_2019}}
        \label{fig5}
        \end{center}
        \end{figure}
       
\vspace{3pt}

       \textbf{Median filters}: The median filter examines each pixel in the image and evaluates its surroundings to determine if it accurately represents its neighboring pixels. It replaces the pixel value with the median value of the neighboring pixels. The median is calculated as the middle value in the sorted list of neighbor pixel values.
       \begin{figure}
       \includegraphics[width=1.1\linewidth]{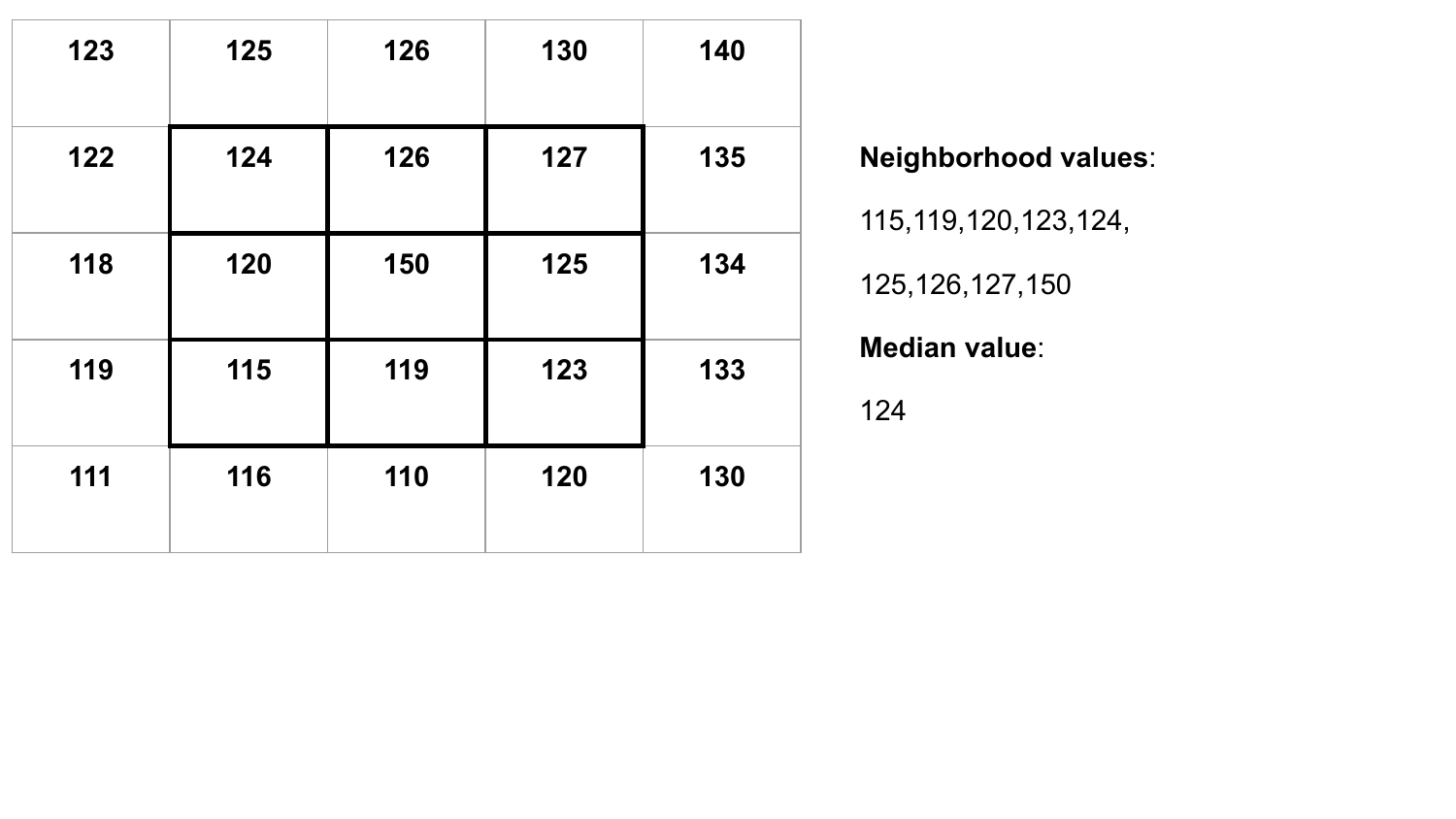}
       \vspace{-55pt}
        \caption{Neighborhood values.\cite{Spatial-Filters-Median-Filter}}
        \label{fig6}
        \end{figure}
        Fig. 6 illustrates the concept of defining neighbors and determining the median value \cite{Spatial-Filters-Median-Filter}. Median filtering is recognized for its robustness against outliers and its ability to better preserve edges and image details compared to mean filtering. However, this method requires additional computational time due to the sorting operation involved.
\vspace{3pt}
        
    \subsubsection{\textbf{Feature extraction}}
The primary objective of the feature extraction unit is to discern and extract pertinent features or structures from the input image data. This step assumes critical significance as it reduces the input data's complexity and directs focus towards the most salient information.

To achieve this, the feature extraction unit employs pre-established algorithms or methodologies that target specific features or patterns within the input image data. These algorithms encompass edge detection, corner detection, blob detection, template matching, and various other techniques. The outcome of the feature extraction unit is a collection of numerical or symbolic descriptors that effectively represent the relevant features or structures existing in the image.numerical or symbolic descriptors that represent the relevant features or structures present in the image.

    \begin{enumerate}
       \item[2.1] {\textbf{Corner and edge detection}}: Corner and edge detection algorithms are fundamental techniques in image processing that aim to identify corners or boundaries between regions with different characteristics in an image. There are many corner and edge detection algorithms, some of the most commonly used ones include:
       \vspace{3pt}

       \textbf{Shi-Tomasi Algorithm}: The Shi-Tomasi algorithm was utilized for feature detection. The Algorithm is employed to identify notable and distinctive points within an image, which can serve as feature points for subsequent analysis or manipulation. It is also a minor modification of the Harris corner detector method, involving the calculation of the score R. In the context of pavement crack detection tasks, the algorithm identifies all interesting and distinctive points in the provided images and detects feature points associated with potential crack locations. This is achieved by examining significant changes in image gradients \cite{infrastructures6080115}.
       \vspace{3pt}

        \textbf{Sobel operator}:The Sobel operator is well-suited for highlighting crack types.  This operator enhances edges by convolving the image with a small kernel (typically 3x3 or 5x5) that calculates the gradient magnitude at each pixel location. It comprises two kernels, one for detecting vertical edges and the other for detecting horizontal edges. Prior to applying the Sobel operator to the pavement crack detection task, it is advisable to preprocess the images by enhancing contrast and removing noise. Once the preprocessing is completed, applying the Sobel operator can effectively detect edges and potential crack locations within the provided images \cite{10.1007/978-981-19-3505-3_11}.
       \vspace{3pt}
 
         \textbf{Canny operator}: The Canny edge detection method is adopted to accurately extract the cracks. The algorithm is an edge detection operator using a multi-stage algorithm to detect a wide range of image edges. The Canny Edge algorithm involves several sequential processes applied to the images. The first step is Gaussian smoothing, where a Gaussian filter is utilized to reduce noise and eliminate small details that are not considered as edges. Subsequently, the Canny algorithm employs Sobel operators to compute the gradient of image intensity and suppress the calculated gradient, retaining only the local maximum values. Following this, two threshold values are used to differentiate between weak edges and strong edges detected by the Sobel operator \cite{app122010651}.
         \vspace{3pt}

         \textbf{Robert operator}: The Robert operator serves as an edge detector that performs a fast and straightforward 2-D spatial gradient measurement on an image. This operator applies a kernel to each pixel in the image, generating a new image where each pixel represents the gradient magnitude at that particular location. However, research conducted by Qingbo et al. in 2016 suggests that the Roberts edge detection operator may produce thicker edges and inaccurately locate them \cite{7783490}.
         \vspace{3pt}

         \textbf{Prewitt operator}: Prewitt operator demonstrates a high accuracy in identifying image edges by utilizing two 3-by-3 matrices known as gradient kernels that are convolved with the image pixels \cite{article}. This operator does not assign any additional weight to the pixels that are closer to the center of the mask. Prewitt operator is favored over Roberts operator when using a larger size kernel as it exhibits better robustness to noise and can detect edges with a higher degree of accuracy \cite{article}.
      \vspace{3pt}
   
       \item[2.2] {\textbf{Image binarization}}: 
       Image binarization involves converting a grayscale or color image into a binary image, where each pixel is assigned a value of either 0 or 1. This process is achieved through thresholding, where each pixel's value is compared to a specified threshold value. Based on this comparison, the pixel is assigned a binary value (0 or 1) depending on whether it is above or below the threshold. Various methods exist for selecting the appropriate threshold value for binarization.
       \vspace{3pt}

       \textbf{Global thresholding (e.g., Otsu)}: Otsu's thresholding are used to determine the threshold value for image binarization \cite{8400199}.  These algorithms calculate a threshold value that separates the image into two classes: foreground (cracks) and background (pavement surface). This algorithm aims to calculate the threshold that minimizes the intra-class variance (i.e., the variance within each class) and maximizes the inter-class variance (i.e., the variance between the two classes). The output of Otsu's method is a binary result segmented into the crack and non-crack regions, enabling further analysis or classification approaches to be applied.
       \vspace{3pt}

        \textbf{Local thresholding}: Local thresholding algorithms are employed to accommodate variations in brightness, contrast, and texture within localized regions of an image by determining optimal thresholds. Localized thresholding divides the image into smaller blocks and calculate a local threshold for each block, enabling the identification of crack pixels within the block using the respective threshold \cite{Katakam2009PavementCD}. The threshold value for each pixel is determined based on the intensity values of its neighboring pixels. Local thresholding methods are particularly beneficial in pavement crack detection tasks as they can adapt to variations in surface texture and lighting conditions, which can affect the visibility of cracks in an image.
\vspace{3pt}

         \textbf{Dynamic thresholding}: Dynamic thresholding is a technique employed to identify dark pixels in images, as these pixels often correspond to potential crack pixels \cite{7077805}. Dynamic threshold methods determine a unique threshold value for each pixel in the image. Initially, a threshold value is calculated for an initial region of the image, and subsequent regions have their thresholds adjusted based on local image properties, such as the mean or median intensity value. The result threshold value is used to segment the region into crack and non-crack regions. Similar to other segmentation approaches, the identified crack and non-crack regions can be further analyzed or processed for subsequent analysis.
      \vspace{3pt}
   
       \item[2.3] {\textbf{Image filtering}}: The objective of image filtering is to enhance specific features or characteristics within an image. Various types of filters can be utilized to achieve this, depending on the particular features that require extraction or enhancement.
       \vspace{3pt}

       \textbf{Gabor filters}: A Gabor filter is a type of linear filter utilized in various image processing applications, particularly for edge detection. It can be conceptualized as a sinusoidal signal with a specific frequency and orientation, modulated by a Gaussian wave. The Gabor filter has shown significant potential in multidirectional crack detection, a task that was previously not explored extensively using this filter \cite{6728529}. The Gabor function possesses a directional property that is sensitive to edges and other features in a particular direction. By convolving an image with multiple Gabor filters having different frequencies and orientations, it becomes possible to extract features from the image that capture its texture, shape, and structure. When applied to an image, a Gabor filter produces the strongest response at edges and points where texture changes occur. Upon subjecting the image to each filter in the filter bank, the detected edge of the circle corresponds to the edge orientation aligned with the orientation of the Gabor filter \cite{MedA}. This observation is visually evident in Fig. 7.
       \begin{figure}
          \includegraphics[width=\linewidth]{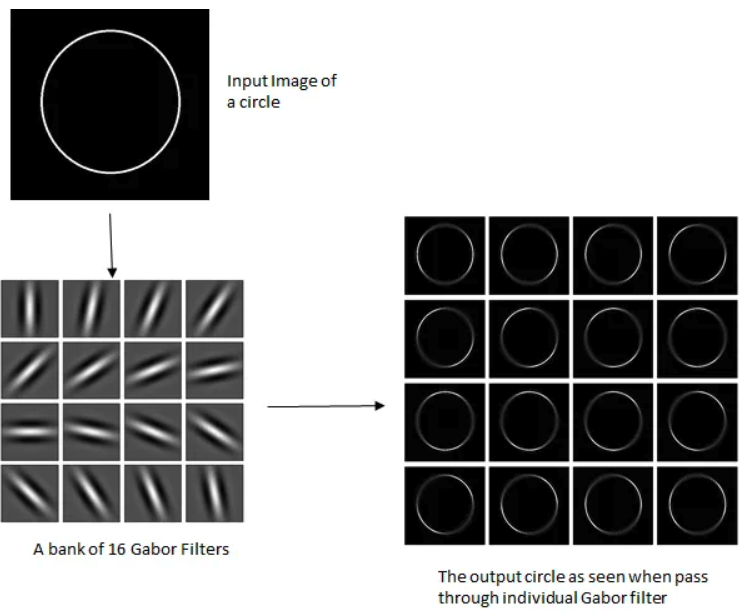}
          \caption{ The corresponding oriented edge features are detected when passed through individual-oriented Gabor filters\cite{MedA}}
          \label{fig7}
        \end{figure}

        \vspace{3pt}

       \textbf{Laplacian of Gaussian Filters (LoG)}: The Laplacian of Gaussian (LoG) filters are used as derivative filters to detect areas of rapid change, such as edges, in images \cite{LGF}. Due to their nature as derivative filters, LoG filters are sensitive to noise and require a smoothing process prior to applying the filters. In crack detection tasks, applying the Laplacian of Gaussian filter in the spatial domain has been shown to be the fastest, most accurate, and most precise method, resulting in the finest detectable crack width \cite{infrastructures4020019}. LoG filter can be applied to the preprocessed image to highlight areas where the intensity changes rapidly, which corresponds to potential crack locations. 
       \vspace{3pt}

       \item[2.4] {\textbf{Image Transform}}: The goal of image transforms is to convert an image from its spatial domain into a different domain, allowing for different types of analysis and processing. In addition to the well-known Fourier transform, there are several other transforms that are utilized in pavement distress detection and classification domain.
       \vspace{3pt}

       \textbf{Wavelet transform}:
       In the Wavelet transform, an image is transformed into a series of coefficients representing the image content at different scales and frequencies (Fig. 8). Wavelets are functions that concentrate on time and frequency around a specific point. The wavelet coefficients capture the details and structure of the image at various scales, enabling the extraction of features such as edges, texture, and patterns. The parameters of the transform can be adjusted to capture both high-frequency and low-frequency information. A separable 2D continuous wavelet transform for several scales is performed for pavement crack detection tasks by Peggy Subirats et al \cite{4107210}. Different wavelet functions possess distinct properties, including smoothing and sharpening effects. The selection of the appropriate wavelet function should be determined by the specific requirements of the image processing task.
       \begin{figure}
          \includegraphics[width=\linewidth]{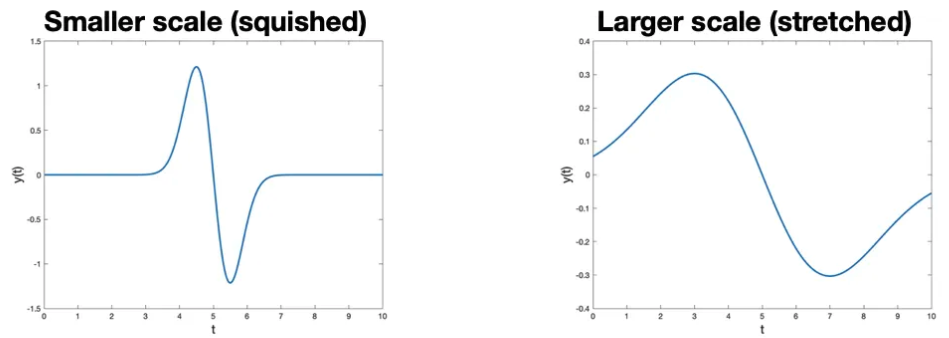}
          \caption{Left: Example wavelet with decreased scale. Right: Example wavelet with increased scale. Image by the author \cite{42c}.}
          \label{fig8}
        \end{figure}

       \vspace{3pt}
 
      \textbf{Beamlet transform}:
      Beamlet Transform is used for the extraction of edges, ridges, and curvilinear objects in digital images. Beamlet transform is considered as a novel tool for high dimensional singularity analysis, as demonstrated by Na Wei et al \cite{5522600}. Being one type of wavelet transform, the Beamlet Transform is a multiscale and multidirectional technique capable of extracting crucial features from images. Through the decomposition of the image into various scales and orientations, it effectively captures both local and global features of cracks. Once the crack features have been extracted using the beamlet transform, a thresholding operation can be applied to eliminate noise and isolate the crack regions within the image. Subsequently, post-processing steps including edge detection, morphological operations, and region growth can be employed to refine the extracted crack results. The effectiveness of the beamlet transform is influenced by various factors, such as the design of the beamlet filter, the number of scales and orientations utilized, and the specific thresholding operation employed. Therefore, it is crucial for the user to select these parameters carefully.
      \vspace{3pt}

      \textbf{Hough transform}:
      The Hough transform can be used to isolate features of a particular shape within an image, which can subsequently be employed as distinguishing features for various computer vision tasks such as image segmentation and object recognition. The Hough transform maps every point in an image to a parameter space, where lines and shapes can be easily detected. Within the parameter space, the lines and shapes manifest as peaks, enabling the extraction of the parameters that define these entities within the original image. Essentially, the Hough transform facilitates the conversion of the image space into the Hough space, wherein shapes in the image space are represented as points in the Hough space. For instance, in the case of line detection, the Hough transform converts lines in the image space into points in the Hough space. The efficacy of the Hough transform algorithm was assessed by Matarneh et al. in the context of detecting and classifying a substantial dataset of highway images, yielding an impressive accuracy rate of 92.14\% \cite{466963dfbe124e44b3ce616ec4b457fb}. In the image space, the equation of a line is typically represented as y = mx + c, where m denotes the slope and c represents the y-intercept of the line. Through the Hough transform, this line equation is transformed into a point (m, c) within the Hough space (as illustrated in Fig. 9). The Hough space provides a distinct representation where lines in the image space to be translated into points, allowing for efficient line detection and characterization.
       \begin{figure}[htbp]
          \includegraphics[width=\linewidth]{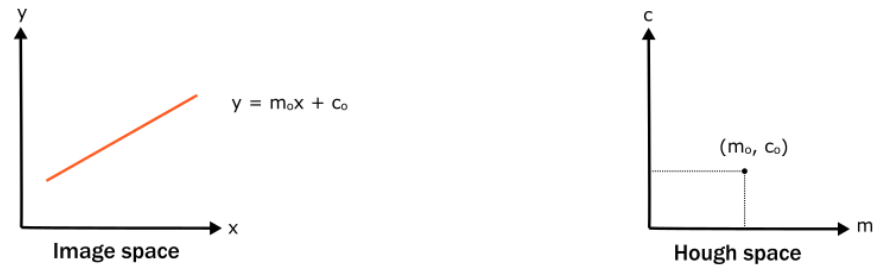}
          \caption{The figure describes the transformation from a line to a Hough space \cite{45c}.}
          \label{fig9}
        \end{figure}

       \vspace{3pt}

     \item[2.5] {\textbf{Feature representation}}: The role of an image representation unit is to extract useful features or representations from an input image that can be used for subsequent tasks, such as object detection, recognition, or segmentation. These features are essentially a condensed version of the original image, capturing the most relevant and informative aspects of the image, and are usually represented by a vector. Some types of image representation methods include:
     \vspace{3pt}

       \textbf{Histogram}:
       A histogram is a graphical representation of the distribution of data.  In the context of image processing, an image histogram provides a graphical representation of the distribution of pixel intensities in a digital image \cite{46c}. For object recognition tasks, a histogram of oriented gradients (HOG) can be calculated to capture the distribution of edge orientations within an image. This information can be useful for identifying the presence and location of objects within the image. Histograms can also be utilized to extract texture information from an image. The frequency and distribution of local image features, such as edges or patterns, can effectively characterize the texture. Additionally, histograms can be employed to compute statistical measures, such as the gray-level co-occurrence matrix (GLCM), which captures important information about the texture properties of an image.
       \vspace{3pt}

       \textbf{Gray-Level Co-occurrence Matrix (GLCM)}:
       The Gray-Level Co-occurrence Matrix (GLCM) is a matrix that captures the distribution of co-occurring pixel values, whether they are grayscale values or colors, at a specific offset within an image. It is a statistical method that examines the texture of an image by considering the spatial relationship of pixels. The GLCM is constructed by counting the occurrences of each pair of gray-level values at the specified offset distance and orientation. The GLCM effectively represents the texture of an image by capturing the spatial dependencies between pixels with different gray levels. It provides a visual representation of the image's texture properties. In the context of image feature extraction, the GLCM features can be used to describe an image's texture in a concise and discriminative manner. By converting texture information into numerical values, the GLCM features can be easily compared and processed using machine learning algorithms. Pramestya et al. utilized the Gray-Level Co-occurrence Matrix (GLCM) as a feature extraction technique to classify various types of road defects, such as potholes, cracks, and other defects. The GLCM features were employed to capture and characterize the texture properties of the road defects, enabling their classification and detection \cite{8534769}.
       \vspace{3pt}

       \textbf{Moments}:
       An image moment refers to a weighted average or function of the image pixel intensities, chosen to have specific properties or interpretations. It is calculated by considering the spatial distribution of pixel intensities in an image. Different types of moments can be calculated, including central, raw, and normalized moments. Central moments are computed with respect to the center of mass of an object, providing information about its shape and orientation, commonly used in object recognition and shape analysis tasks. Raw moments are calculated with respect to the origin of the coordinate system and describe the overall distribution of pixel intensities in an image. They can be utilized for feature extraction purposes. Chou et al. \cite{399871} conducted pavement distress classification using Hu moments, Zernike moments, and Bamieh moments. These moment-based features were employed to accurately classify pavement distress, achieving a reported classification accuracy of one hundred percent.
       \vspace{3pt}

       \textbf{Principal Component Analysis (PCA)}:
       After the feature extraction process, a feature reduction unit can be applied to reduce the dimensionality of the feature vector. Principal Component Analysis (PCA) is a widely used method for feature reduction \cite{8047878}. PCA aims to retain the maximum amount of dataset variation while reducing the feature dimensionality. Abdel-Qader et al. \cite{ABDELQADER2006771} developed three PCA-based algorithms for crack classification in images. These algorithms included the original PCA method, PCA with linear features extracted from the raw data, and PCA with features extracted from local areas. Among these methods, PCA utilizing local area characteristics demonstrated the highest accuracy in crack detection. 
       \end{enumerate}
       \vspace{3pt}

    \subsubsection{\textbf{Classification}}
    The role of the classification unit is to assign labels or categories to input images based on their extracted features or representations. This task is typically accomplished using a classifier, which is a machine learning algorithm trained on labeled data to learn the relationship between features and corresponding labels. Supervised and unsupervised algorithms are two primary types of machine learning algorithms being used and trained as classifiers. Each type has its own characteristics and applications.
       \vspace{3pt}

   \begin{enumerate}
    \item[3.1] {\textbf{Supervised}}:
      Supervised learning proves valuable when dealing with a labeled dataset of pavement crack images, where each image is annotated with precise information concerning crack size and type. This labeled data serves as a training resource for machine learning models, particularly convolutional neural networks (CNNs), enabling them to effectively identify and differentiate various types of cracks in pavement images. Following model training, the CNN can then be applied to classify new pavement images, accurately categorizing them based on the presence and distinctive characteristics of cracks.
        \vspace{3pt}

        \textbf{Multilayer perceptron (MLP) neural network}:  
        An MLP is a feedforward neural network consisting of multiple layers of nodes or neurons. In the context of pavement crack classification, an MLP can be trained to learn the patterns and characteristics of different types of cracks. 
        To utilize an MLP for pavement crack classification, a substantial dataset of labeled crack images is typically used to train the network. Images are fed into the input layer of the network, enabling the network to recognize patterns in the images associated with different types of cracks. The hidden layers of the network perform nonlinear transformations on the input data, enabling the model to acquire more intricate representations of the crack images. Finally, the output layer of the network produces a classification decision based on the learned features.
        \vspace{3pt}

        \textbf{Support Vector Machine (SVM)}:
        Support Vector Machine (SVM) is a classic kernel model utilized for binary and multi-class classification tasks. SVM can separate irregularly distributed data into binary or multiple classes by determining the optimal hyperplane that effectively separates the dataset based on its features. The application of SVM has demonstrated efficacy in distress detection and classification tasks. According to \cite{SHOLEVAR2022104190}, SVM achieves an accuracy of 78\% in recognizing three types of cracks.
        \vspace{3pt}

        \textbf{Least Squares Support Vector Machine (LSSVM)}:
        LS-SVM (Least Squares Support Vector Machine) differs from traditional SVM by solving for the optimal linear function that maps the input data to the output labels, rather than focusing on maximizing the margin between classes. It achieves this by minimizing the sum of squared errors between the predicted output labels and the actual output labels, subject to constraints that ensure the classifier is as accurate as possible. The use of kernel functions in LS-SVM allows it to handle non-linearly separable data by mapping it into a higher-dimensional space where linear separation becomes possible. In the study result provided by \cite{article51}, LS-SVM demonstrated remarkable performance, achieving the highest classification accuracy rate of approximately 89\% and an impressive area under the curve value of 0.96. This underscores its effectiveness in pavement crack detection tasks, particularly when combined with feature extraction techniques that can effectively capture the essential characteristics of pavement cracks.
        \vspace{3pt}

        \textbf{Naïve Bayesian Classifier (NBC)}:
        NBC (Naive Bayes Classifier) calculates the probability of each class by analyzing the training data and then determines the likelihood of the features belonging to each class. By applying Bayes' theorem, these two probabilities are combined to calculate the posterior probability of each class given the features. The class with the highest posterior probability is then selected as the predicted class for a given instance. NBC is known for its computational efficiency and ability to handle a high degree of feature dependencies, including disjunctive and conjunctive concepts \cite{domingos1997optimality}.
        \vspace{3pt}

        \textbf{Decision Tree (DT) and Random Forest(RF)}:
        The decision tree is a classifier algorithm based on a tree structure. The algorithm classifies the input dataset into binary or multiple branches at each stage along the path from the root node to leaf nodes (terminal nodes) based on features. At each leaf node, predictions of a category or a numerical value are made. The decision-making process of a decision tree is guided by evaluating the entropy, which measures the impurity of the data. The algorithm aims to maximize the decrease in entropy at each stage, resulting in branches that exhibit greater purity. Consequently, the data is classified with greater clarity and distinction.CrackForest \cite{7471507} is a crack detection model structured based on Random Structured Forests (RSF), which utilizes the principles of decision trees in its framework.
        \vspace{3pt}

         \textbf{K-Nearest Neighbor (KNN)}:
         The K-nearest neighbors (KNN) algorithm is a non-parametric supervised classifier that relies on a pre-labeled dataset. The algorithm predicts the label of unlabeled data by considering the K data points closest to the unlabeled data in the feature space. By determining the majority class among the K nearest neighbors, the algorithm assigns a label to the unlabeled data point. In the context of crack detection, \cite{article54} presented an example of enhancing the basic KNN model by incorporating Differential Evolution (DE) to optimize its performance. This hybrid approach aims to improve the accuracy and effectiveness of crack detection.
         \vspace{3pt}

         \textbf{AdaBoost}:
         AdaBoost (Adaptive Boosting) is a prevalent boosting technique that combines multiple weak classifiers to build one robust classifier \cite{55c}.  In the context of pavement crack detection, AdaBoost can be applied by training the algorithm on a labeled dataset consisting of images categorized as either containing a crack or not. 
        During the training process, AdaBoost learns a set of weak classifiers, each specialized in identifying specific features within the image that are indicative of a crack. For example, one weak classifier may focus on detecting straight lines, while another may specialize in identifying irregular lines. Each weak classifier assigns a score based on the presence of its identified features. AdaBoost combines the scores from the weak classifiers to generate a final score, which is used to make the ultimate decision regarding whether the image contains a crack or not. 
        \vspace{3pt}

        \textbf{Gradient boosting (GB) and gradient-boosted decision trees (GBDT)}:
        Gradient boosting is a boosting algorithm that optimizes models using the gradient descent algorithm \cite{10.1145/3357384.3358149}. Gradient boosting is a boosting algorithm that optimizes models using the gradient descent algorithm \cite{10.1145/3357384.3358149}. It operates by iteratively fitting new predictors in the direction of the gradient descent of the previous predictor and aiming to minimize the residual errors made by the previous predictor. In Gradient Boosting Decision Trees (GBDT), decision trees are used as the base learners. During the training process, each decision tree is built to minimize the residual errors made by the previous decision tree. Each new decision tree focuses on capturing the patterns and information that were not captured by the previous trees. By iteratively adding decision trees and adjusting their predictions based on the gradient descent direction, the overall model gradually improves its ability to predict the target variable.
        \vspace{3pt}

        \textbf{Fuzzy Classification Method(FCM)}:
        FCM (Fuzzy C-means) is an algorithm that assigns linguistic values, such as "high," "low," and "medium," to each feature of input images, representing the degree of similarity with fuzzy sets \cite{ZHANG2018350}. This process, known as fuzzification, allows for the handling of uncertainty and imprecision in the input data. FCM is especially advantageous in pavement crack detection scenarios where cracks may be partially obscured or exhibit irregular shapes.
        \vspace{3pt}

         \textbf{Minimum Cost Spanning Tree(MCST)}:
         MCST (Minimum Cost Spanning Tree) constructs a connected and undirected graph with minimal edge weights. In a pavement crack classification task, MCST takes a set of crack images and creates a graph, where each node represents one crack image and an edge represents the similarity between pairs of images. The weights on these edges are determined by examining crack image features such as texture, color, and shape \cite{10.1007/978-3-030-80316-2_1}. By finding the minimum spanning tree of the generated graph, MCST identifies the most critical images in the dataset and classifies them based on the features used to construct the graph. This can is particularly beneficial in pavement crack classification tasks, as it enables the algorithm to focus on the most informative and significant features of the images \cite{9204935}.
         \vspace{3pt}

      \item[3.2] {\textbf{Unsupervised}}:
      On the other hand, unsupervised learning techniques can be used to identify patterns and structures in unlabeled data, such as clustering algorithms. In pavement crack detection, unsupervised learning can be used to group similar types of cracks based on their visual features without prior knowledge of the specific types of cracks present in the dataset.
      \vspace{3pt}

        \textbf{K-Means Clustering}:  
        K-Means is an unsupervised clustering algorithm used for classification tasks. For K clusters, K data points are chosen as centroids, and the rest data points are assigned to different clusters. The distance between each data point to the closest centroid is being calculated. The selection and assign process is repeated until optimized centroids are found. K-Means algorithm has been applied in crack classification \cite{10.1007/978-3-030-80316-2_1} and segmentation tasks \cite{9204935}.
        \vspace{3pt}

        \textbf{Gaussian mixture models(GMM)}:
        GMM (Gaussian Mixture Model) is an unsupervised machine-learning technique used for clustering and density estimation. It involves multiple Gaussian Distributions, with each representing a cluster. The GMM algorithm consists of three main stages: Initialization, Expectation, and Maximization. In the Initialization stage, the number of clusters or Gaussian distributions and their parameters are randomly assigned. In the Expectation stage, the probability of each data point belonging to each Gaussian distribution is computed using Bayes' theorem. The Maximization step utilizes the computed probabilities as weights to re-estimate the parameters of each Gaussian distribution. The expectation and maximization stages are repeated until convergence is achieved \cite{8998713}.
      \end{enumerate}
\vspace{3pt}

      \subsubsection{\textbf{Segmentation}}
      Segmentation algorithms are valuable in addressing the pavement distress problem as they enable the automatic identification and separation of various distress types within pavement images.  These algorithms are capable of distinguishing between different types of cracking among other distress types. Fig. 10 illustrates several techniques utilized for image segmentation, which are discussed below:
     \begin{figure*}[t]
     \begin{center}
          \includegraphics[width=0.9\textwidth]{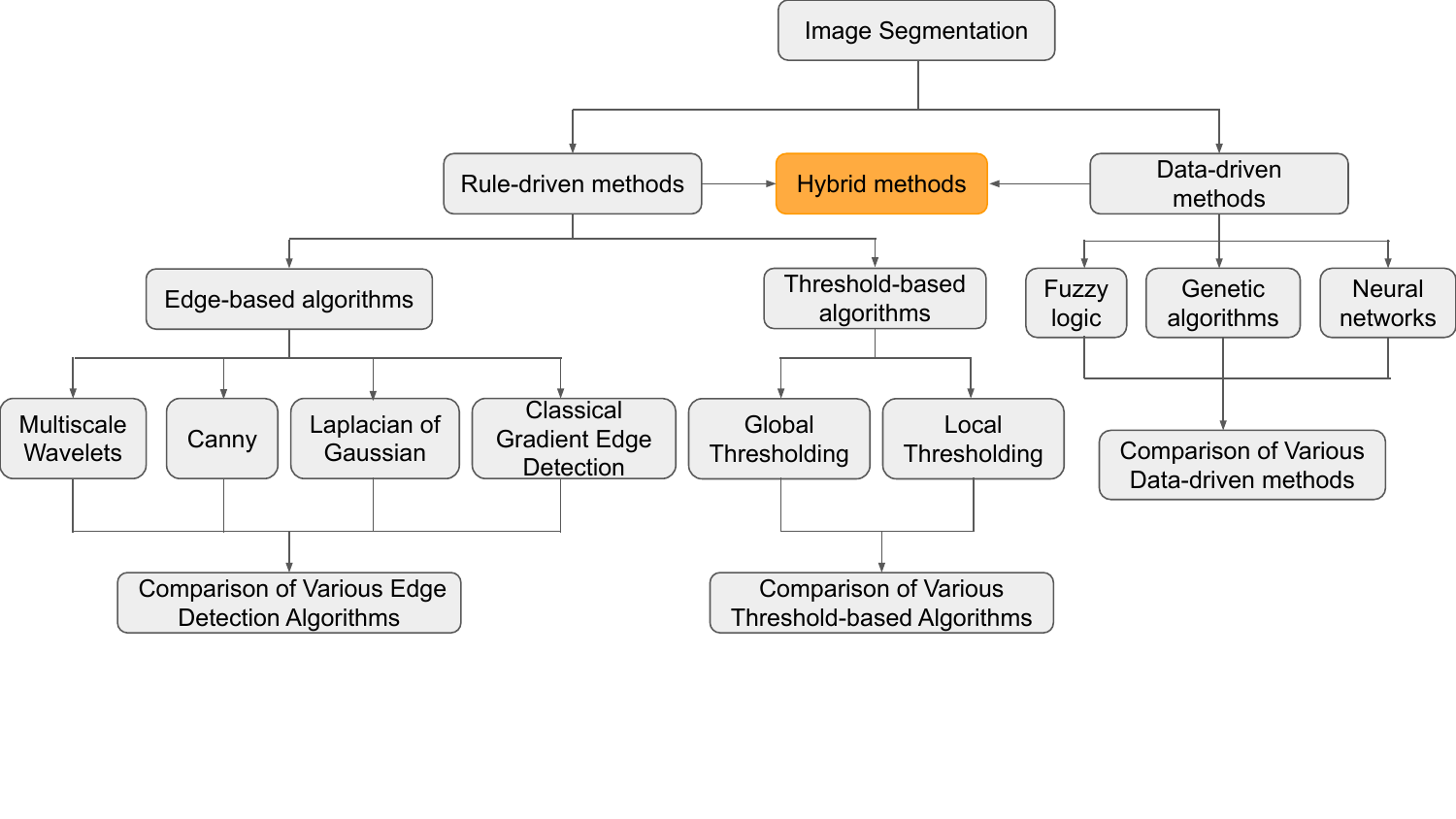}
          \vspace{-50pt}
          \caption{The image-based techniques for pavement distress segmentation.\cite{KHERADMANDI2022126162}}
          \label{fig10}
        \end{center}
        \end{figure*}
 \vspace{3pt}
       
    \begin{enumerate}
      \item[4.1] {\textbf{Rule-driven methods}}:
      Image segmentation could be processed based on edge and threshold by applying rules. The edge-based algorithms operate on the principle that changes in light, color, shadow, and texture result in the formation of edges in digital images. These algorithms focus on detecting linear features corresponding to crack boundaries \cite{KHERADMANDI2022126162}. Four commonly used edge detection techniques are Classical Gradient, Laplacian of Gaussian, Canny edge detection, and Multiscale Wavelet \cite{KHERADMANDI2022126162}. The threshold technology categorized pixels into objects and backgrounds based on the gray scale of pixels. It assumes that the grayscale of the background remains consistent and that distresses appear darker than the background. However, determining the appropriate threshold value can be challenging due to the non-uniform frequency distribution of gray-level values in actual pavement images. Global thresholding assumes that the grayscale frequency distribution of the image can be represented by a bimodal histogram. As a result, a single threshold value is sufficient to separate the object from the background. On the other hand, local thresholding divides the image into smaller regions and determines a threshold value for each sub-image individually. In the process of thresholding, smoothing filters such as the Gaussian filter, extremum sharpening, and adaptive filtering can be employed to assist in determining the threshold value effectively. 
  \vspace{3pt}
    
      \item[4.2] {\textbf{Data-driven methods}}:
      The development of automotive image segmentation heavily relies on the availability and quality of datasets. Fuzzy logic is a methodology used to train or learn from fuzzy image datasets. It focuses more on the application aspect rather than relying solely on the dataset itself. This approach offers significant advantages when dealing with regions that cannot be precisely defined. Genetic algorithms are utilized to enhance the quality of images and efficiently perform segmentation, especially when confronted with a large search space and limited objective function information \cite{Bradshaw_2020}.In the realm of neural networks, various architectures have been applied for image segmentation in the pavement distress domain. These include Convolutional Neural Network (CNN), Deep Convolutional Neural Networks (DCNNs), Region-based CNN (R-CNN), YOLO (You Only Look Once), and other pre-trained deep convolutional neural networks (PDCNN) such as AlexNet, ResNet, SqueezeNet, GoogleNet, Xception, Visual Geometry Group (VGG), ShuffleNet, MobileNetV2, HRNet, and more. Several segmentation neural networks have been proposed, including U-Net, DAUNet (Deep Augmented UNet), Bi-Directional Cascade Network (BDCN), Holistically Nested Edge Detection (HED), Richer Convolutional Features for Edge Detection (RCF), CrackU-net, Crack Transformer (CT) \cite{GUO2023104646}, DeepLabv3 \cite{chen2017rethinking}, PSPNet \cite{8100143}, DeepLabV3+, CGnet \cite{9292449}, W-segnet \cite{ZHONG2022104436}, and others. These deep learning networks demonstrate the ability to automatically segment certain types of pavement distress with acceptable accuracy.
     \end{enumerate}
     
\subsection{\textbf{Learning-based algorithms}}
Various algorithms have been applied in the field of crack detection and identification, with a focus on learning-based approaches. These methods encompass the utilization of single models as well as combinations of multiple algorithms. Previous research has conducted comprehensive reviews of the application of machine learning methods in pavement condition detection, providing both general and detailed descriptions of their significance \cite{SHOLEVAR2022104190} \cite{kim2017concrete}.

In Section IV.A, it was discussed that the traditional approach to distress detection and identification primarily relies on Support Vector Machine (SVM), Decision Tree (DT), and Random Forest (RF) algorithms. While these traditional machine learning methods are capable of performing the task, their performance is limited due to their lower generalization ability.

To overcome these limitations, Deep Neural Networks (DNN) architectures have emerged as a promising solution. These networks have demonstrated superior performance by automatically extracting and learning features from input images. The following sections introduce different types of DNNs categorized based on their network architecture.
\vspace{3pt}

    \subsubsection{\textbf{Feedforward networks}}
    
    Feedforward networks have a series of layers that process input data in a linear manner without feedback connections.
    \vspace{3pt}

    \begin{enumerate}
        \item [1.1] {\textbf{LeNet-5}}:
        LeNet-5 is a pre-trained feed-forward convolutional neural network that serves as the foundation for various image-processing network structures, such as AlexNet, VGG, and ResNet. LeNet-5 consists of seven layers, including two sets of one convolutional layer followed by one subsampling layer, two fully connected layers, and one classification layer. The convolutional layer includes convolution, pooling, and nonlinear activation functions to extract features from input images.  A modified LeNet-5 network, as demonstrated by Qu, Z. \cite{9040553}, has been successfully employed for crack detection and classification tasks, showcasing its accurate performance in solving these tasks.
        \vspace{3pt}

        \item [1.2] {\textbf{Fuzzy Artificial Neural Networks(FANNs)}}:
        A fuzzy neural network (FNN) can be conceptualized as a three-layer feedforward network. It consists of a fuzzy input layer responsible for fuzzification, a hidden layer that incorporates fuzzy rules, and a final fuzzy output layer responsible for defuzzification \cite{CHAUDHRY2007111}. The neurons within FNNs make decisions based on fuzzy logic, allowing them to handle inputs with varying lighting conditions and image quality. This characteristic enables FNNs to effectively deal with uncertainty and imprecision in the input data.
        \vspace{3pt}

        \item [1.3] {\textbf{Probabilistic ANNs (PNN)}}:
        The probabilistic neural network (PNN) is constructed based on the principles of the Bayes theorem. The network architecture consists of an input layer that receives the image features, a pattern layer that computes probability distributions of the input features, and a summation layer that combines the probability distributions. Finally, an output layer produces the classification results \cite{9827988}. PNN is particularly advantageous in handling noisy data. In the field of crack classification, PNN has been recognized as a useful algorithm, especially in the early stages of its development \cite{9827988}.
        \vspace{3pt}

        \item [1.4] {\textbf{Radial Basis Function Neural Network (RBFNN)}}:
        The radial basis function neural network (RBFNN) is an artificial neural network that comprises a single hidden layer. In this network, radial basis functions are utilized as activation functions, which measure the distance between an evaluated point and the center of each neuron to compute the corresponding weights. The weights decrease as the distance between the evaluated point, and a neuron increases. Improved versions of RBFNN and complex-valued RBFNN, have been employed in crack detection tasks, particularly in the early stages of research, to identify the location and depth of cracks \cite{machavaram2013identification}.
        \vspace{3pt}

        \item [1.5] {\textbf{RetinaNet}}:
        RetinaNet is a detection model that consists of a backbone network and two sub-networks \cite{ale2018road}. The backbone network of the RetinaNet, known as the Feature Pyramid Network (FPN), is responsible for extracting feature maps. The two sub-networks, namely the classification subnets and the box regression subnets, perform the tasks of classification and box regression, respectively. RetinaNet, the loss function employed, combines the focal loss for classification and the smooth L1 loss for regression. This choice of loss function is motivated by the recognition that one-stage classification and regression models can be influenced by imbalanced datasets. The focal loss effectively addresses this issue by assigning greater weight to classes that are more challenging to classify, thereby enhancing the overall performance of the model.
    \end{enumerate}
    \vspace{3pt}

    \subsubsection{\textbf{Convolutional neural networks (CNNs)}}
    Convolutional neural networks (CNNs) are specialized for processing image data and use convolutional layers to extract features from the input.
    \vspace{3pt}

    \begin{enumerate}
        \item [2.1] {\textbf{Fully Convolutional Network (FCN)}}:
        FCN, also known as Fully Convolutional Network, is a neural network that consists of convolutional layers and a pooling layer. In contrast to Convolutional Neural Networks, FCN does not include fully-connected layers, resulting in a reduced parameter count. The output of FCN is a dense prediction, wherein the network predicts a semantic segmentation image at the pixel level. In the domain of pavement distress detection, Yang et al. \cite{yang2018automatic} utilized the FCN network architecture, which facilitated an end-to-end structure capable of not only learning to identify distress based on local features but also determining the location of distress by incorporating global features.
        \vspace{3pt}

        \item [2.2] {\textbf{Convolutional Neural Networks (CNN) and DCNN}}:
        Convolutional Neural Networks (CNNs) are neural networks that employ convolutional and pooling layers. Convolutional layers extract and learn features by applying convolution filters to scan images and generate feature maps containing block-level features. Pooling layers, on the other hand, downsample feature maps to reduce dimensionality. CNN has emerged as the predominant approach for image-based analysis \cite{mostafa2021review}. With the advancement of machine learning algorithms, researchers have begun to explore the possibility of combining multiple algorithms and models to enhance distress detection performance. In their study, Junzhi Z. et al. \cite{zhai2022automatic} integrated a channel attention algorithm with the CNN architecture, aiming to enhance the features extracted by CNN and thereby improve efficiency and detection accuracy.
        \vspace{3pt}

        \item [2.3] {\textbf{Visual Geometry Group (VGG)}}:
        The Visual Geometry Group (VGG) is a traditional CNN architecture that has been widely used in computer vision tasks. In VGG, the convolutional layers extract feature maps by utilizing a kernel size of 3 x 3, while the pooling layers employ a filter with a size of 2 x 2. By augmenting the depth of CNN networks, the VGG architecture effectively enhances their performance. Notably, VGG-16, comprising 16 hidden layers, and VGG-19, consisting of 19 hidden layers, have emerged as two prominent variations. In a comprehensive study conducted by Wafae H. et al. \cite{hammouch2022crack}, a comparative analysis was conducted to evaluate the performance of CNN and VGG-19 networks in tasks pertaining to crack classification and detection. Furthermore, Zhengchao et al. developed a VGG-16-based pixel-level crack detection network in their research \cite{pang2020research}.
        \vspace{3pt}

        \item [2.4] {\textbf{AlexNet}}:
        AlexNet is a substantial convolutional neural network built upon the architectural principles of LeNet-5. The network excels in processing RGB images with a fixed batch size of 227 x 227 x 3 \cite{dung2019autonomous}. Diverging from LeNet-5, AlexNet boasts a more profound structure, comprising five convolutional layers and three fully connected layers. While the activation function has been transitioned from tanh to ReLU, the pooling methodology remains as average pooling. To enhance its generalization capacity, AlexNet integrates dropout layers following the last pooling layer and the first fully connected layer.
        \vspace{3pt}

        \item [2.5] {\textbf{ZFNet}}:
        ZFNet possesses the capability to visualize the features within the intermediate layers of CNN networks \cite{zhai2022automatic}. It bears a similarity in structure to AlexNet, albeit with a diminished number of filters. ZFNet comprises eight convolutional layers, three fully connected layers, and unpooling layers that facilitate the restoration of data from a pooled state. Additionally, it accomplishes an approximate inverse by precisely recording the locations of the maxima within each pooling region.
        \vspace{3pt}

        \item [2.6] {\textbf{ResNet, Res2Net and ResNesT}}:
        With the increasing depth of CNN networks, there is a potential decline in their performance due to random gradient descent. To tackle this issue, a solution called ResNet was proposed. ResNet introduces a residual block and offers two mapping methods: identity mapping and residual mapping. The fundamental concept is to highlight small changes occurring between convolutional layers while disregarding redundant information during the optimization process. Ullah et al. \cite{dung2019autonomous} conducted a comparative analysis of the performance of AlexNet, ResNet18, and SqueezeNet on an 8-class road crack dataset for crack detection tasks. In a recent study, Fan Z. et al. \cite{hammouch2022crack} employed a parallel ResNet structure for pavement distress detection and demonstrated the model's accuracy using the CFD and CrackTree200 datasets. Another variation, Res2Net, replaces the filter within bottleneck residual blocks with a series of smaller filters to enhance feature information at multiple scales. ResNesT is a ResNet variant that incorporates a Split-Attention network, consisting of split-attention blocks, to capture cross-feature interactions.
        \vspace{3pt}

        \item [2.7] {\textbf{Inception-ResNet-v2}}:
        The Inception-ResNet-v2 architecture integrates residual blocks and inception modules from the Inception and ResNet architectures. The inclusion of residual blocks enables the network to learn residual mappings between layers, thereby improving both accuracy and efficiency. Concurrently, the incorporation of inception modules allows the network to capture multi-scale features by utilizing parallel convolutional filters of different sizes. As a result, Inception-ResNet-v2 presents a compelling approach to enhance the capabilities of detection networks and accurately define crack areas \cite{pang2020research}.
        \vspace{3pt}

        \item [2.8] {\textbf{InceptionV3}}:
        The InceptionV3 network \cite{dung2019autonomous} is a pretrained network that integrates multiple convolutional, pooling, and fully connected layers. It incorporates a unique module called the "Inception module," which is specifically designed to capture information at various scales by utilizing parallel convolutional filters with different sizes. The Inception module consists of four parallel convolutional pathways that are later merged at the output. Each pathway contains convolutional filters of different sizes that operate simultaneously on the input feature maps. These filters include sizes of 1 x 1, 3 x 3, 5 x 5, and a max pooling operation. In recent times, InceptionV3 has exhibited remarkable performance in pavement detection tasks, primarily attributed to its exceptional ability to handle image classification tasks.
        \vspace{3pt}

        \item [2.9] {\textbf{U-Net}}:
        U-Net is a convolutional encoder-decoder network that comprises a contracting path and an expansive path. The contracting path conducts downsampling to extract features, while the expansive path performs upsampling to combine the feature maps extracted from the downsampling process with newly generated feature maps. This combination process effectively recovers edge feature loss that occurs during downsampling. U-Net has been successfully utilized for identifying both single classes and multiple classes of pavement distress \cite{li2021pixel,yu2022improved,chen2020automatic,cheng2018pixel,jiang2020deep}.
        \vspace{3pt}

        \item [2.10] {\textbf{SegNet}}:
        SegNet is a convolutional encoder-decoder network. The encoder comprises a 13-layer network identical to the first 13 layers of the VGG-16 networks. To reduce the number of parameters and maintain the resolution of input images, the encoder structure eliminates the FCN layer \cite{CHEN2020100144}. On the other hand, the decoder upsamples the feature map to carry out pixel classifications. SegNet shares similarities with U-Net in terms of structure. However, a notable distinction is that U-Net forwards entire feature maps to the decoder, while SegNet utilizes max-pooling indexes to selectively perform upsampling on specific regions of the feature map.
        \vspace{3pt}

        \item [2.11] {\textbf{SPP-Net}}:
        SPP-Net is a convolutional neural network utilized for object detection and classification. By substituting the last pooling layer before the FCN with a Spatial Pyramid Pooling (SPP) layer, SPP-Net effectively reduces computational costs \cite{deng2020concrete}. In the RCNN approach, region proposals were resized into fixed-size images, which were then processed by the CNN to generate feature maps. To mitigate the impacts resulting from the resizing process, the SPP layer employs various scale poolings and combines the resulting feature maps as input for the CNN.
        \vspace{3pt}

        \item [2.12] {\textbf{DenseNet}}:
        DenseNet is a convolutional neural network with dense connections. In this architecture, every two layers within a dense block are directly connected, and each layer within a dense block receives the feature maps from all preceding layers. To control the size of features, transition blocks comprising convolution layers and pooling layers are used to connect the dense blocks. A specific variant, DenseNet 201, is composed of 201 deep layers and has been utilized for crack detection on the CFD dataset \cite{mei2020densely}.
        \vspace{3pt}

        \item [2.13] {\textbf{GoogLeNet}}:
        GoogLeNet is a deep CNN-based architecture. In contrast to VGG, GoogLeNet introduced the Inception model, which employs convolution with three different kernel sizes: 5 x 5, 3 x 3, and 1 x 1 \cite{ni2019pixel}. The resulting feature map is a fusion of these three feature maps. GoogLeNet reduces training time by employing a multi-path structure and utilizing 1x1 convolutional layers to regulate the number of input and output channels.
        \vspace{3pt}

        \item [2.14] {\textbf{Lookup-based Convolutional Neural Network and Part-based classification network}}:
        The Lookup-based Convolutional Neural Network (LCNN) is a neural network that conducts convolution by convolving the input with all the vectors in the dictionary, eliminating the need to reconstruct the weight tensor. This algorithm aims to optimize the training process of a traditional CNN model by reducing effort. Chen et al. \cite{chen2020pothole} have introduced a novel structure for pothole detection by utilizing the LCNN network for distress detection and a part-based classification network (PCNN) for binary distress classification. The LCNN identifies candidate regions, while the PCNN performs binary classification to distinguish between positive and negative regions within the candidates.
        \vspace{3pt}

        \item [2.15] {\textbf{SqueezeNet}}:
         SqueezeNet is a lightweight CNN model designed to achieve comparable accuracy on image classification tasks with significantly fewer parameters compared to AlexNet. This reduction in parameters is accomplished by replacing 3x3 convolution filters with 1x1 filters \cite{ullah2021comparative} and reducing the number of input channels for the filters. SqueezeNet employs Fire modules, which comprise a squeeze layer and an expanded layer. The squeeze layer utilizes 1x1 convolutions to decrease the number of input channels, while the expansion layer combines features using both 1x1 and 3x3 convolution layers.
         \vspace{3pt}

         \item [2.16] {\textbf{ShuffleNet}}:
         Group convolution is a convolution technique that involves separating channels into groups and performing convolution operations on each group separately. ShuffleNet introduces channel shuffling, which allows channels within each group to contain feature information from other channel groups. The efficacy of the channel shuffle technique has been demonstrated. Qayyum et al. \cite{qayyum2023assessment} conducted a performance comparison of GoogLeNet, MobileNet-V2, Inception-V3, ResNet-18, ResNet-50, ResNet-101, and ShuffleNet for pavement crack detection and classification tasks.
         \vspace{3pt}

         \item [2.17] {\textbf{You Only Look Once (YOLO)}}:
         YOLO is a real-time object detection algorithm that combines region proposal selection with object detection. The main distinction between YOLO and RNN is that the network performs a single pass over the image. The model divides the image into small regions and identifies regions containing objects using bounding boxes. The YOLO series, which includes Tiny YOLO and YOLO V1 to V5, has been utilized for crack detection. In comparison to RNN and Faster RCNN, YOLO can be trained in a shorter time period. Keyou G. et al. \cite{guo2022pavement} optimized YOLO V5 by incorporating the attention mechanism, CIOU algorithm, and K-means algorithm to enhance the performance of YOLO V5 in pavement distress identification.
         \vspace{3pt}

         \item [2.18] {\textbf{Darknet}}:
         VGG, ResNet, and Darknet are commonly employed as backbones for object detectors \cite{pan2022spatiotemporal}. Darknet is a convolutional neural network (CNN) known for its simplicity and efficiency. It consists of multiple layers of convolutional filters and pooling operations, followed by fully connected layers that produce the final predictions. Darknet incorporates "skip connections," which connect lower-level layers to higher-level layers within the network. These "skip connections" enable the network to capture both low-level features (such as edges and corners) and high-level features (such as object shapes and textures). 
         \vspace{3pt}

         \item [2.19] {\textbf{Shot Multibox Detector (SSD)}}:
         The Shot Multi-Box Detector (SSD) is a precise object detection algorithm that utilizes the VGG-16 convolutional neural network as its foundation. Like the YOLO network, SSD operates using a single forward pass architecture. SSD proves to be particularly adept at detecting small objects, as it overcomes a limitation present in the YOLO network, where cells are constrained to possess only one label and two bounding boxes \cite{8534769}.
         \vspace{3pt}

        \item [2.20] {\textbf{Xception and MobileNet}}:
        Xception is an innovative architecture that combines depth-wise separable convolution with Inception models, resulting in a distinct separation of the convolution stage into two steps: depthwise convolution and pointwise convolution \cite{doychevadecentralised}. This approach is built on the premise that cross-channel correlation and spatial correlation can be treated independently. During the depthwise convolution step, each convolution filter exclusively operates on one channel, disregarding cross-channel information. Subsequently, a 1x1 convolution is applied to the combined feature map, capturing cross-channel information through pointwise convolution. MobileNet, inspired by Xception, also adopts the depthwise convolution followed by a pointwise convolution structure \cite{10.1007/978-981-19-2445-3_3}. Furthermore, MobileNet further reduces the number of parameters to enhance overall performance.
        \vspace{3pt}

        \item [2.21] {\textbf{ASINVOS and ASINVOS-mod net}}:
        ASINVOS Net is a convolutional neural network (CNN) model utilized for distress detection, comprising four 2D convolutional layers, three max-pooling layers, and three fully connected layers \cite{8560372}. To enhance its performance, the ASINVOS-mod Net, a modified version of ASINVOS Net, introduced alterations to the filter sizes and substituted the max-pooling layers with reducing convolutional layers Eisenbach et al. \cite{7966101}. conducted a comparative analysis of the ASINVOS Net and ASINVOS-mod Net on the German Asphalt Pavement Distress Dataset (GAPs) for crack classification. The study demonstrated that both models exhibited high accuracy in performing the task. 
    \end{enumerate}

    \vspace{3pt}

    \subsubsection{\textbf{Recurrent neural networks (RNNs)}}
    Recurrent neural networks (RNNs) are specialized in processing sequential data and utilize recurrent layers to preserve the memory of previous inputs.
    \vspace{3pt}

    \begin{enumerate}
        \item [3.1] {\textbf{Region-Based Convolutional Neural Network (RCNN), Faster R-CNN, and Mask R-CNN}}:
        Based on convolutional neural networks (CNNs), RCNN focuses on object detection and semantic segmentation by incorporating region proposals. The architecture can be structured as a segmentation process that generates region proposals, which represent regions that potentially contain objects. These proposals undergo a feature extraction process conducted by CNN and a classification process utilizing the SVM algorithm. Notably, RCNN enhances efficiency by performing the convolution and pooling processes on the region proposals instead of the entire image. To further improve efficiency, Faster R-CNN restructures RCNN into an end-to-end process. In this approach, the original image is fed into the CNN network, and the resulting feature map is used to obtain region proposals, which are then resized using the ROI pooling layer. The classification stage is accomplished through the softmax layer. However, during the process of resizing regions of interest (ROIs), there is a potential for location and size bias in the bounding boxes of the region proposals compared to the original ones. To address this issue, the ROI Align process is introduced in Faster R-CNN, leading to the development of a new RCNN model known as Mask R-CNN. Xu, X. et al. \cite{s22031215} conducted a comparative study between Faster R-CNN and Mask R-CNN using various types of crack images. The results indicate that both models demonstrate excellent performance in detecting single cracks.
        \vspace{3pt}

        \item [3.2] {\textbf{Cascade-RCNN}}:
        Cascade-RCNN \cite{ren2023preprocessing} is a multi-stage RCNN model developed to address the challenges associated with increasing IoU (Intersection over Union) thresholds in the RCNN model. In each stage of the cascade, a detector is trained using distinct IoU thresholds, and the output from the previous stage serves as input for the subsequent stage. This approach enables each stage's detector to concentrate on detecting region proposals that fall within specific IoU thresholds. Tong, Z. et al. employed the Cascade-RCNN \cite{TONG201869} framework to achieve automated detection of pavement cracks.
        \vspace{3pt}

        \item [3.3] {\textbf{Dynamic-RCNN}}:
        Dynamic-RCNN \cite{ren2023preprocessing} is a two-stage object detection algorithm that dynamically adjusts the label assignment criteria and loss function based on the training features of a given detection task. The Dynamic Label Assignment (DLA) technique is employed to adapt the label assignment criteria by dynamically raising the IoU threshold, thereby enhancing the quality of proposals. As the quality of positive region proposals improves, the influence of these proposals is reduced based on the definition of Smooth L1 Loss, which in turn limits the model's performance. Consequently, the loss function requires adjustment. Therefore, The Dynamic Smooth L1 Loss (DSL) approach is utilized to modify the loss function.
        \vspace{3pt}

        \item [3.4] {\textbf{Grid-RCNN}}:
        Grid-RCNN is a two-stage algorithm that shares the RPN network with the RCNN network. However, Grid-RCNN and RCNN differ in their prediction methodologies. Grid-RCNN utilizes FCN to predict the location of grid points, which are points that exhibit inner spatial correlation and enable the differentiation between objects and the background. The boundaries of region proposals are then determined based on the predicted grid points. In the context of the automatic pavement crack localization task, Ren, R. et al. \cite{ren2023preprocessing} compared Cascade-RCNN, Dynamic-RCNN, and Grid-RCNN, all employing ResNet50-FPN as the backbone.
        \vspace{3pt}

        \item [3.5] {\textbf{Region proposal network (RPN)}}:
        The Region Proposal Network (RPN) is a prominent target detection algorithm that revolves around the concept of regional proposals \cite{8981093}. It constitutes an end-to-end network that generates region proposals by leveraging feature maps. By employing a sliding window approach, the RPN scans through feature maps derived from the CNN, thereby generating anchors with two potential labels: foreground or background. The RPN network serves as a foundational structure for numerous region proposal models, such as RCNN and Faster R-CNN \cite{8981093}.
        \vspace{3pt}

        \item [3.6] {\textbf{Elongated pavement distress Region-based Convolutional Neural Network}}:
        The Elongated pavement distress Region-based Convolutional Neural Network (Epd RCNN), developed by Huiqing X. et al. \cite{xu2022elongated}, represents an enhanced version of the RCNN network specifically tailored for detecting elongated pavement distress. Diverging from the traditional RCNN approach, the Epd RCNN combines distinct levels of features to mitigate the loss of feature information. This algorithm incorporates Multi-Scales and Aspect-Ratios Mechanisms to generate region proposals that closely align with the cracks. After the ROI pooling layer, a space attention module is employed to reinforce foreground features. In comparison to Faster RCNN and Faster RCNN+, the Epd RCNN demonstrates superior performance in crack detection tasks with reduced foreground information \cite{xu2022elongated}.
        \vspace{3pt}

        \item [3.7] {\textbf{Neural Architecture Search Network (Nasnet)}}:
        Neural Architecture Search (NAS) is a framework based on the Reinforcement Learning Network (RNN), where an RNN controller is employed to acquire and generate hyperparameters for the model. In the context of the Neural Architecture Search Network (Nasnet), an RNN controller is utilized to predict hyperparameters and introduces the concept of blocks and cells. These cells can be classified as normal or reduced cells, with normal cells employed when the input and output feature resolutions remain the same, and reduction cells utilized when the input feature resolution is half that of the output features. For image classification tasks, NASNet-Mobile represents a pretrained model trained on the ImageNet database. König et al. \cite{doychevadecentralised} conducted a comparative analysis of Inception V3, MobileNet V1, and NASNet for pavement distress detection and classification tasks, evaluating their individual performances.
        \end{enumerate}

 \vspace{3pt}
       
    \subsubsection {\textbf{Generative adversarial networks (GANs)}}
    
    GANs comprise a generator and a discriminator, with the generator aimed at generating data that closely resembles the actual dataset, while the discriminator's goal is to discern the authenticity of the input data. In their work, Liu Z. et al. \cite{LIU2023104674} employed various lightweight GAN models for distress classification. The utilization of these lightweight GANs resulted in improved accuracy and efficiency in the classification of four distinct distress classes, surpassing the performance of ResNet without GAN \cite{LIU2023104674}.
    
\section{Public datasets}
For the training and validation of machine learning algorithms, previous studies have employed privately collected datasets in combination with public datasets to ensure an unbiased assessment of performance. Notably, the focus of distress detection and classification research has predominantly been on crack-related phenomena. According to the findings of Munawar, H. S. et al. \cite{infrastructures6080115}, a substantial 70\% of the reviewed papers in \cite{infrastructures6080115} utilized privately collected datasets obtained from various devices. Conversely, only 30\% of the research relied on public datasets. The reviewed papers commonly employed the following public datasets for validation and training purposes: Crack500, GAPs384, CrackTree200, CFD, Aigle-RN, CrackTree260, CrackWH100, CrackNJ156, CFD-ex, and CrackForest.

\begin{table*}[h!]
  \begin{center}
    \caption{A detailed view of public datasets}
    \label{tab:table1}
    \begin{tabular}{p{4cm}|c|c|p{4cm}|c|r} 
      \textbf{Dataset Name} & \textbf{Distress type} & \textbf{Dataset size} & \textbf{Image pixel size}& \textbf{Reference}& \textbf{Aerial}\\
      \hline
      Crack500 & Crack & 500 & 2000 x 1500 & Yang et al. \cite{yang2019feature} & No\\
      CFD & Crack & 118 & 480 x 320 & Shi et al. \cite{7471507} & No\\
      AigleRN & Crack & 38 & 460 × 990 & Chambon et al. \cite{soldovieri2011noninvasive}  & No\\
      GAPs384 & Crack & 1969 & 1920 x 1080 & Eisenbach et al. \cite{eisenbach2017get} &No\\
      CrackTree200 & Crack&206&800 x 600&Zou et al. \cite{ZOU2012227}&No\\
      CrackTree260 &Crack &26&512 × 512&Zou et al. \cite{ZOU2012227}&No\\
      CrackWH100 &  Crack &100&512 x 512&Zou et al. \cite{8517148}&No\\
      CrackNJ156 &Crack&156&512 x 512&Xu et al. \cite{XU2022102825}&No\\
      DeepCrack &Crack&537&544 × 384&Liu et al. \cite{LIU2019139}&-\\
      Road Damage Dataset (RDD2022)&Crack and pothole&47,420& 512x512, 600x600, 720x720, 3650x2044 & Arya et al. \cite{9377790}&Included\\
      Cracks-and-Potholes-in-Road-Images-Dataset & Crack and pothole&2235&At least 1280 x 729 &Passos et al. \cite{passos2020cracks}&No\\
      GAPs 10m&Multiple  distresses&20&5030 x 11505 &Stricker et al. \cite{9551591}&No\\
      GAPs v1&Multiple distresses&1969&1920 x 1080&Eisenbach et al. \cite{7966101}&No\\
      GAPs v2 &Multiple distresses&2468  &1920 x 1080& Stricker et al. \cite{8852257}&No\\
      UAPD &Multiple distresses&2401&512 x 512&Zhu et al. \cite{ZHONG2022104436}&Yes
    \end{tabular}
  \end{center}
\end{table*}

\section{System evaluation}

The condition of road pavements is intricately linked to the ride quality experienced by road users. Early research conducted by Carey and Irick in 1960 laid the groundwork for investigating pavement performance and its correlation with serviceability \cite{SHOLEVAR2022104190}. The present serviceability index (PSI), introduced as a measure of pavement performance, typically diminishes with the aging of pavements, reflecting the pavement's quality. However, it is essential to acknowledge that the PSI index solely accounts for the pavement's robustness factor, overlooking other influential factors, including environmental conditions and human intervention.

In addition to the PSI, the Federal Highway Administration (FHWA) in the USA introduced the pavement condition index (PCI) as an alternative metric to assess the long-term performance of pavements. The PCI incorporates measurement procedures for each type of distress, considering their severity and extent. It employs a rating scale ranging from 0 to 100 to predict pavement conditions. Well-established charts based on field climatic conditions and empirical knowledge are consulted to calculate the standard PCI. However, it is crucial to recognize that the variability of conditions across different regions can lead to inaccurate predictions without proper calibration.

The International Roughness Index (IRI) is a mathematical parameter employed to characterize the two-dimensional profile of a road. Various measurement methods can be utilized to compute the IRI, encompassing traditional static rods, level surveying equipment, and advanced high-speed inertial profiling systems \cite{peraka2020pavement}.

The Structural Number (SN) denotes the comprehensive structural capacity necessary to sustain the specified traffic loads in the pavement design. It serves as an abstract value that quantifies the pavement's structural strength, considering factors such as soil support (MR), cumulative traffic represented by equivalent single axle loads (ESALs), terminal serviceability, and environmental conditions \cite{doi:10.1061/JPEODX.0000373}.

Skid Resistance (SR) is defined as the longitudinal coefficient of friction (COF) between a wetted road surface and a standard reference material. Its determination is conducted using a standardized friction tester at a velocity of 70 km/h \cite{zakeri2017image}.

Various metrics are commonly utilized during the evaluation of image-based learning models, including Precision, Recall, Accuracy, and F1-score. These metrics enable quantitative assessment of the detection performance of machine vision-based methods for analyzing pavement distress. Precision indicates the ratio of correctly predicted instances to the total number of predictions made by the model. Conversely, Recall measures the proportion of correctly detected instances out of all instances that should have been identified. The F1-score, representing the harmonic mean of precision and recall, ranges from 0 to 1, where 0 denotes the poorest detection performance and 1 signifies the best. The expressions for precision, recall, and F1-score are provided below.
$$ \text{Precision} = \frac{TP}{TP + FP}$$
$$ \text{Recall} = \frac{TP}{TP + FN}$$
$$ \text{F1-score} = \frac{2 * Precision * Recall }{Precision + Recall}$$
\vspace{3pt}

TP represents a positive result correctly detected by the method, TN denotes a negative result correctly detected by the method, FP indicates a negative example incorrectly detected as a positive example, and FN refers to a positive example that is either incorrectly detected or remains undetected.
In the evaluation of object detection tasks, such as crack detection, multiple researchers utilize Intersection over Union (IoU), mean IoU, and Average IoU metrics. IoU is a value ranging from 0 to 1 that quantifies the extent of overlap between the predicted bounding box and the ground truth bounding box. An IoU of 0 indicates no overlap between the boxes, while an IoU of 1 implies that the union of the boxes is equal to their overlap, indicating complete overlap. The mean IoU (MIoU) score measures the average accuracy of segmentation across different classes. The expressions for IoU and MIoU are presented below.

$$ IoU = \frac{Prediction Region \cap Ground Truth}{Prediction Region \cup Ground Truth}$$

$$ MIoU = \frac{IoU}{n}$$

\section{Evaluation results}
To assess and improve the accuracy and efficiency of pavement distress detection, evaluating various algorithms is of utmost importance. In this section, a comparison of evaluation metrics for different pavement distress detection algorithms is provided. The comparison is presented in Table 2.

\begin{table*}[h!]
\renewcommand{\arraystretch}{1.5} 
  \begin{center}
    \caption{Result table (The research conducted by Zhang and Yuen (2021) \cite{zhang2021crack} is based on aerial images)}
    \label{tab:table2}
    \begin{tabular}{p{3.5cm}|p{4cm}|p{2cm}|p{6cm}} 
      \textbf{References} & \textbf{Dataset} & \textbf{Base method} & \textbf{Metrics}\\
      \hline
      Zhang et al. (2016)\cite{7533052} & Local Dataset & ConvNets& Precision = 0.8696, Recall = 0.9251\\
      &&& and F1-score = 0.8965\\
      \hline
      Eisenbach et al. (2017)\cite{7966101} & German Asphalt Pavement Distress(GAPs)(Public) &CNN&Accuracy = 0.9136\\
      Zhang et al. (2017)\cite{ZHANG2017130} &Local Dataset &Pixel-SVM Shadow modeling
      & Precision = 0.9013, Recall = 0.8763\\
      &&&and F1-score=0.8886\\
      Pauly et al. (2017)\cite{10.22260/ISARC2017/0066}&Local Dataset &CNN& Accuracy = 0.913; Precision = 0.907; \\
      &&&Recall =0.920\\
      \hline
      Tong et al. (2018)\cite{doi:10.1080/14680629.2017.1308265} &Local Dataset &AlexNet/DCNN &Accuracy = 0.9436 and MSE = 0.2377\\
      Nguyen et al. (2018)\cite{10.1145/3287921.3287949} &JTIRC Dataset (Local)&CNN&Precision = 0.91, Recall = 0.891\\ 
      &CrackIT Dataset (Public) & & and F1-score = 0.90\\
      Fan et al. (2018)\cite{fan2018automatic} &CFD (Public) &CNN (Public) &Precision = 0.9119, Recall = 0.9481\\
      & AigleRN Dataset && and F1-score = 0.9244\\
      \hline
Fan et al. (2019) \cite{8814000} & Public dataset by Özgenel \cite{ozgenel2018concrete} 227x227 (40000 images) &k-means clustering &Accuracy = 0.9870\\
Kouzehgar et al. (2019) \cite{KOUZEHGAR2019102959} &Locad Dataset captured by robots
&CNN&Accuracy: 0.8967; Precision: 0.8690 \\
&&& Recall: 0.8180; F1 measure: 0.8420 \\
Park et al. (2019)\cite{doi:10.1061/(ASCE)CP.1943-5487.0000831} & Local Dataset captured by four vehicles & CNN & Accuracy: 0.9045 Precision: 0.9081 \\
& & & Recall: 0.7440 F1 measure: 0.8179 \\
\hline
Wang et al. (2021)\cite{WANG2021103484}& Local Dataset 299 × 299(15,000) & Inception-ResNet & Precision = 0.9837, Recall = 0.9382 \\
& & &and F1-score = 0.9599\\
Zhang and Yuen (2021) \cite{zhang2021crack} &Public dataset by Li et al. 2019 \cite{li2019research}
1024 × 1024(2068) &FF-BLS&Accuracy = 0.9672\\
\hline
Ren et al. (2022)\cite{electronics11213622}&Public and Local Dataset &YOLOV5&  Precision = 0.953; F1-score = 0.889 \\
&&&Recall =0.834\\
Shi et al. (2022)\cite{doi:10.1177/14759217221140976}&Public Datasets&Unet& Accuracy = 0.9932; Precision = 0.9123 \\
&&&Recall = 0.7946; F1-score = 0.8217 \\
\hline
Xiao et al. (2023)\cite{XIAO2023103172}&Public Dataset Crack500 &high-resolution network (HRNet)&  Precision = 0.9154; F1-score = 0.9196 \\
&&&Recall =0.9238\\
Maurya et al. (2023)\cite{doi:10.1080/10298436.2023.2180638}&Public Datasets&Encoder-Decoder& Precision = 0.8413; F1-score = 0.8264\\
&&&Recall = 0.8120\\
Jing et al. (2023)\cite{jing2023multi}&Public Datasets&ResNet& Precision = 0.832; F1-score = 0.821\\
&&&Recall = 0.810\\
    \end{tabular}
  \end{center}
\end{table*}
\vspace{-5pt}

\section{Open-source Codes}
Conducting a comprehensive review of prior research necessitates the collection of open-source codes, which play a pivotal role in the scientific process and contribute to knowledge advancement in specific fields. By openly sharing these codes, new researchers can validate the original work's results and findings, fostering transparency and trust within the scientific community. Moreover, accessible open-source codes encourage collaboration among researchers, enabling them to build upon existing work and explore novel technological avenues. In the context of pavement stress detection tasks, an examination of available open-source codes offers valuable context and insights.
One GitHub user has generously provided an open-source code that showcases an approach leveraging the Keras and TensorFlow libraries alongside ZF, VGG16, GoogleNet, ResNet50, or ResNet101 neural networks. Most of these codes are implemented in Python \cite{public_test}. The implementation process requires Python version 3.6 and the TensorFlow and Keras packages. The dataset utilized in this approach comprises aerial images captured by unmanned aerial vehicles, proving highly pertinent to the proposed methodology and aligning with the researchers' background knowledge. However, opportunities for further performance enhancement when training conventional image processing models like VGG16, ResNet, and GoogleNet with this aerial image dataset have yet to be explored. Furthermore, the user offers an alternative approach incorporating the Fast R-CNN model for object detection, delineated through the following steps:
\begin{enumerate}
\item{Preprocessing: The initial step involves preprocessing the input image, which includes resizing the image to a fixed size and normalizing the pixel values.}
\item{Region Proposal: Fast R-CNN utilizes the Selective Search algorithm to generate regions of interest (RoIs) within the input image that may contain objects. These RoIs are generated based on various visual cues, such as color and texture.}

\item{CNN Feature Extraction: The RoIs are then passed through a pre-trained convolutional neural network (CNN) to extract visual features specific to each RoI. The CNN generates a feature map that represents the visual characteristics of the RoIs.}

\item{RoI Pooling: The feature map is subsequently inputted into an RoI pooling layer, which extracts a fixed-size feature vector from each RoI. This is achieved by dividing each RoI into a predetermined number of sub-regions and performing max pooling on each sub-region. The result is a fixed-size feature vector for each RoI.}
\item{Fully Connected Layers: The fixed-size feature vectors are fed into a series of fully connected layers to classify each RoI into predefined categories and predict the bounding box coordinates for each detected object.
}
\item{Non-maximum Suppression: Finally, the output detections undergo post-processing using non-maximum suppression (NMS) to eliminate redundant detections.}
\end{enumerate}
By combining region proposal and classification within a single neural network, Fast R-CNN simplifies the architecture and reduces computational overhead. The RoI pooling layer allows the network to handle RoIs of varying sizes and aspect ratios, thereby improving object detection accuracy. Furthermore, the inclusion of NMS helps reduce the number of redundant detections and enhances the precision of the model.

\section{Future Research Direction}
Prior investigations have extensively delved into implementing artificial intelligence (AI) and machine learning methodologies for detecting and classifying pavement distress. Review papers have effectively categorized algorithms based on diverse approaches. Nonetheless, certain challenges persist, warranting further exploration. As outlined in this paper, numerous algorithms utilized by researchers have demonstrated commendable accuracy in distress detection and classification. However, conducting equitable comparisons among these methods proves challenging due to the dearth of publicly available pavement distress datasets. Some models are trained and evaluated on private datasets, while others are tailored to address specific distress types within particular scenarios.

Additionally, despite the increasing adoption of unmanned aerial vehicles (UAVs) for image acquisition, publicly accessible datasets predominantly comprise images captured by mobile phones or linear cameras. This distinction between UAV-applied and UAV-not-applied categorizations accentuates a research gap concerning algorithms that explore different image sources. Future efforts should be directed towards creating comprehensive public datasets and exploring additional algorithms for pavement distress detection and classification tasks, leveraging UAV-acquired image datasets. Moreover, potential advancements can be achieved through investigations into end-to-end solutions and integrated approaches to simultaneously address distress detection and classification tasks.
\section{Conclusions}
This paper presents a comprehensive and up-to-date review of state-of-the-art Artificial Intelligence (AI) applications employed for detecting and classifying pavement distresses using 2D images. The integration of AI applications with various semi-automated and automated data collection methodologies has substantially facilitated the identification and categorization of pavement distress, offering time-efficient and cost-effective solutions. From 2018 to 2023, convolutional neural network-based algorithms emerged as the prevailing techniques for effectively detecting and classifying pavement distresses in image data.

Over the past three decades, researchers have actively pursued the detection of pavement cracks, and the utilization of image processing techniques has proved to be a practical approach for pavement distress detection, thereby enhancing road maintenance and safety. Employing methods such as image preprocessing (including filters and non-linear filters), thresholding, feature extraction (such as edge detection operators, image segmentation, and image transformers), and feature representation, automated detection and characterization of pavement cracks have been achieved with remarkable accuracy and efficiency. These techniques offer the potential to reduce the time and cost associated with manual crack detection while yielding more reliable results. Overall, this review underscores the significance and promising applications of image processing techniques in the field of pavement crack detection.

Prior research has established that a range of machine learning and deep learning methodologies, encompassing traditional supervised and unsupervised techniques, as well as feedforward networks, convolutional neural networks, recurrent neural networks, and generative adversarial networks, can attain impressive accuracy levels in the detection and classification of pavement distress. Nonetheless, the existing studies predominantly focus on a limited set of distress types, particularly cracks, and are constrained by the scarcity of publicly available datasets. In recent years, the advent of UAV-based image acquisition coupled with image processing models has emerged as an innovative approach to data collection. However, their implementation in pavement distress analysis remains relatively constrained. Given the growing prevalence of UAV technologies, it is envisaged that UAV-based approaches will exert a more substantial impact and confer significant advantages in pavement distress detection and classification.

 
%

\bibliographystyle{ieeetr}
\bibliography{mylib}

\begin{IEEEbiography}[{\includegraphics[width=1in,height=1.25in,keepaspectratio]{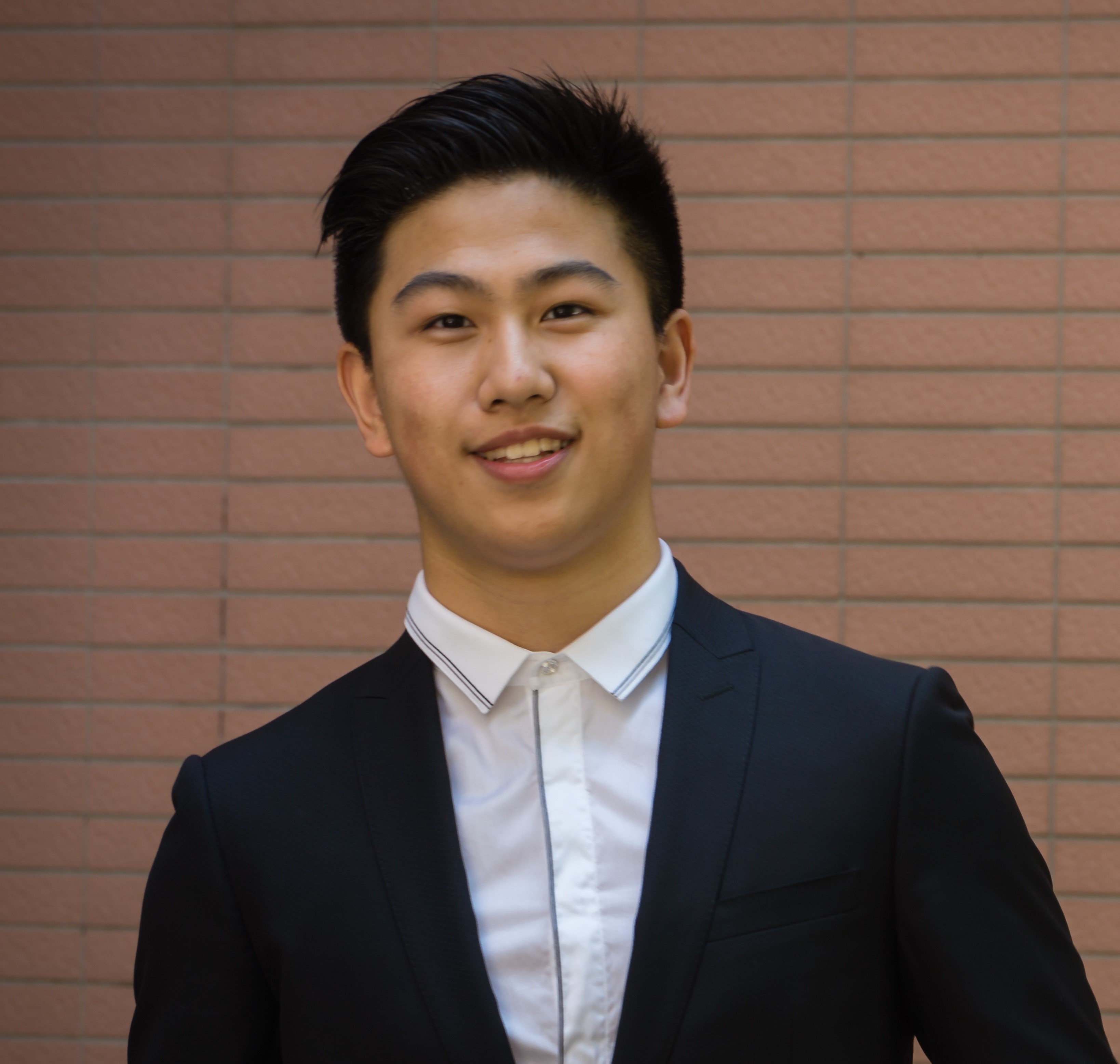}}]%
{Sizhe Guan} is a dedicated student pursuing a Master's degree in the faculty of engineering Systems and Technology at McMaster Univerisity, Hamilton, Canada. Currently focusing on automation and smart system, he is working on an innovative research paper to optimize traffic management using AI and data analytics.
\end{IEEEbiography}

\vspace{-0.5cm}

\begin{IEEEbiography}[{\includegraphics[width=1in,height=1.25in,keepaspectratio]{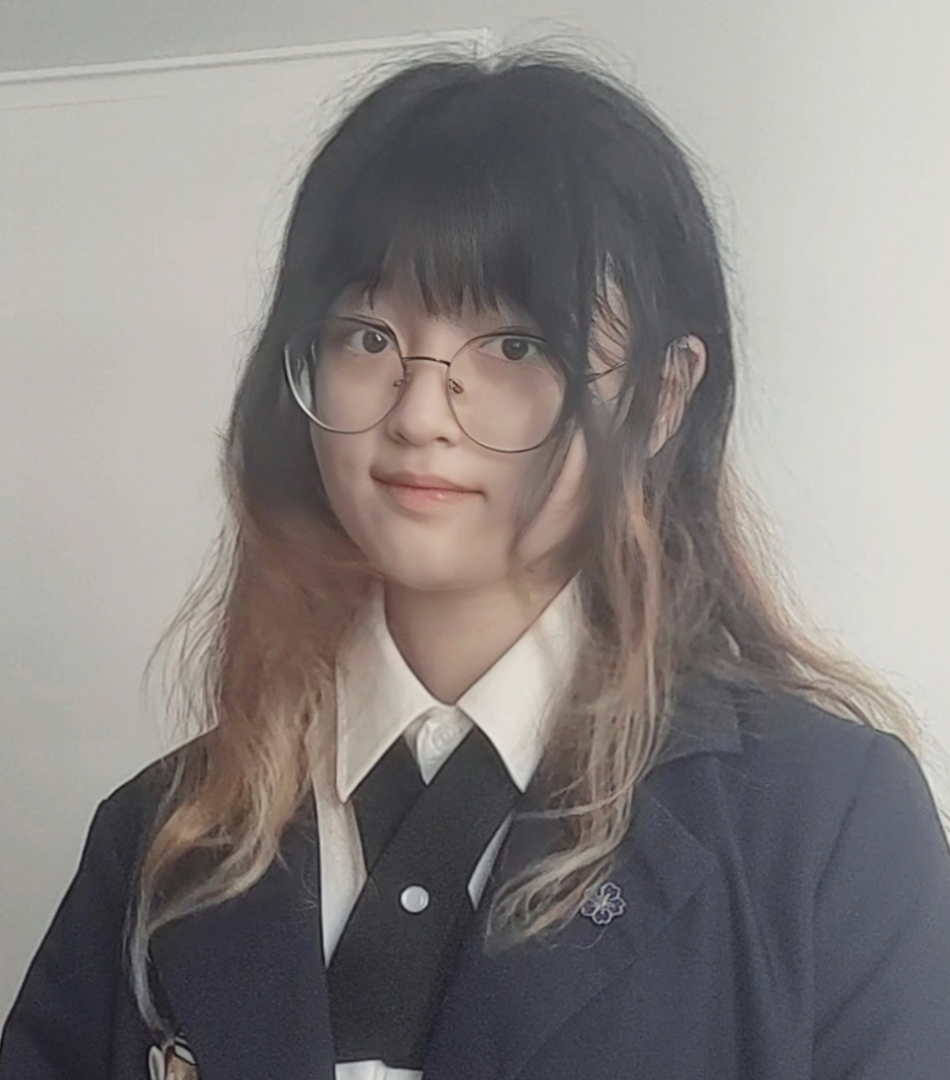}}]%
{Haolan Liu} is a dedicated student pursuing a Master's degree in the faculty of engineering Systems and Technology at McMaster Univerisity, Hamilton, Canada. She is focusing on software development, machine learning, natural language processing. 
\end{IEEEbiography}

\vspace{-0.5cm}

\begin{IEEEbiography}[{\includegraphics[width=1in,height=1.25in,keepaspectratio]{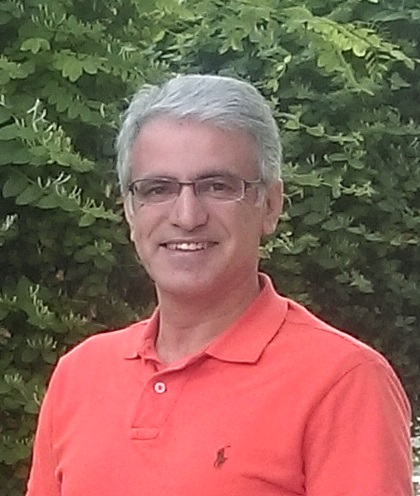}}]%
{Hamid-Reza Pourreza} received the M.Sc. degree in electrical engineering and the Ph.D. degree in computer engineering from Amirkabir University of Technology, Tehran, Iran, in 1993 and 2002, respectively. He is a Professor with the Department of Computer Engineering, Ferdowsi University of Mashhad, Mashhad, Iran. His research interests include image processing, machine vision, and intelligent transportation systems.
\end{IEEEbiography}

\vspace{-0.5cm}

\begin{IEEEbiography}[{\includegraphics[width=1in,height=1.25in,keepaspectratio]{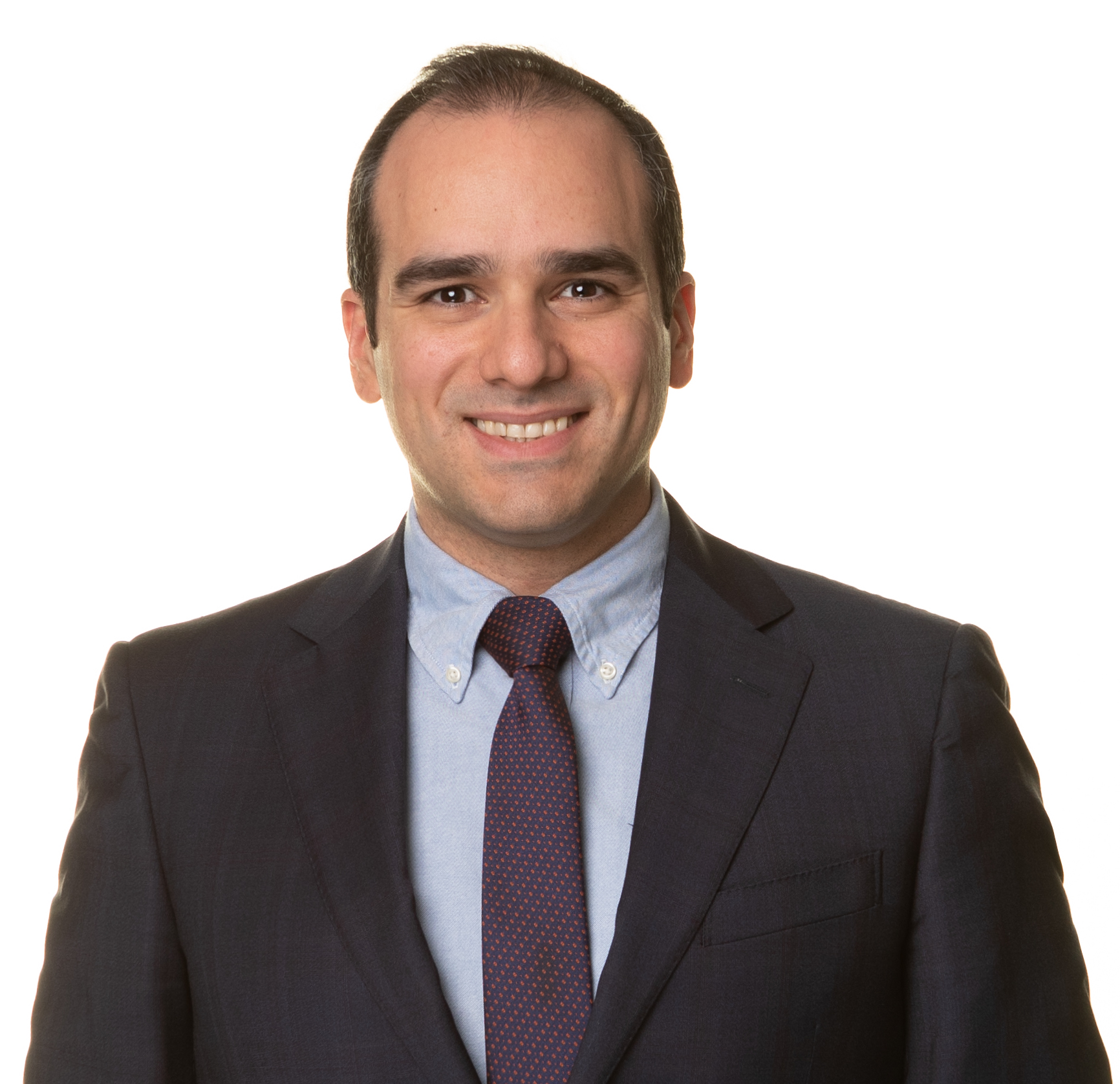}}]%
{Hamidreza Mahyar} is an Assistant Professor in the Faculty of Engineering at McMaster University. Before joining MacMaster, he was a postdoctoral research fellow at Boston University and Technical University of Vienna. Dr. Mahyar received his Ph.D. in Computer Engineering from Sharif University of Technology. His research interests lie at the intersection of machine learning, network science, and natural language processing, with current emphasis on the fast-growing subjects of Generative AI, Large Language Models, and Graph Neural Networks.
\end{IEEEbiography}\vfill

\end{document}